\documentclass[fleqn,10pt]{wlscirep}
\usepackage[utf8]{inputenc}
\usepackage[T1]{fontenc}

\usepackage{array,booktabs}
\newcolumntype{M}[1]{>{\centering\arraybackslash}m{#1}} 
\usepackage{subcaption}
\usepackage{graphicx}
\usepackage{xspace}

\usepackage{multirow}

\newcommand{\monthsRunningDaily}{3\xspace}
\newcommand{\daysRunningDaily}{90\xspace}
\newcommand{\dailyFoundEvents}{43\xspace}

\newcommand{\avgAlertsPerDay}{0.51\xspace}
\newcommand{\avgMinsPerDay}{12.76±0.5\xspace}
\newcommand{\dailyTotalScenes}{7976\xspace}
\newcommand{\dailyHoursExplore}{81.06\xspace}

\newcommand{\totalEMITscenes}{220k\xspace}

\newcommand{\archiveCorrectPrevMissed}{157\xspace}
\newcommand{\archiveHighPriorityAlarms}{646\xspace}

\newcommand{\nhthreeEvents}{78\xspace}
\newcommand{\notwoEvents}{219\xspace}
\newcommand{\coEvents}{51\xspace}

\newcommand{\dnorm}{$D_{norm}$\xspace}
\newcommand{\nhthree}{NH$_{3}$\xspace}
\newcommand{\notwo}{NO$_{2}$\xspace}
\newcommand{\cotwo}{CO$_{2}$\xspace}
\newcommand{\chfour}{CH$_{4}$\xspace}

\newcommand{\zenodoDataset}{\url{https://doi.org/10.5281/zenodo.19992726
}\xspace}
\newcommand{\codeModels}{\url{https://github.com/emit-sds/daily-trace-gases/}\xspace}

\newcommand{\fix}[1]{#1}
\newcommand{\fixS}[1]{}
\newcommand{\fixM}[2]{#1}

\title{Fully Automatic Trace Gas Plume Detection}

\author[1,*]{Vít Růžička}
\author[1]{David~R.~Thompson}
\author[1]{Jay~E.~Fahlen}
\author[1]{Amanda~M.~Lopez}
\author[1]{Steven~Lu}
\author[2]{Chuchu~Xiang}
\author[1]{Holly~Bender}
\author[1]{Daniel~Jensen}
\author[1]{Philip~G.~Brodrick}
\author[1]{Jake~Lee}
\author[1]{Brian~Bue}
\author[3]{Daniel~H.~Cusworth}
\author[4]{Luis~Guanter}
\author[1]{Adam~Chlus}
\author[1]{Andrew~Thorpe}
\author[1]{Robert~O.~Green}
\affil[1]{Jet Propulsion Laboratory, California Institute of Technology, 4800 Oak Grove Drive, Pasadena CA, USA}
\affil[2]{Independent researcher, Houston TX, 77041, USA}
\affil[3]{Carbon Mapper, 680 E Colorado Blvd, Pasadena, CA, 91101, USA}
\affil[4]{Universitat Politècnica de València, Valencia, Spain}

\affil[*]{ruzicka@jpl.nasa.gov}

\begin{abstract}
Future imaging spectrometers will expand contemporary data volumes by orders of magnitude, requiring automated methods to upscale labor-intensive detection of trace gas point sources. Here we present a fully-automated approach that achieves operational performance for plume detection and labelling without human participation.  Our method combines machine learning (ML)-based morphological analysis with physics-based spectroscopic model fitting. We deploy it on data from the EMIT imaging spectrometer, operating in two modes. First, we present a ``daily digest'' that runs automatically on all downlinked data, flagging the largest events for immediate response. The daily digest demonstrates that a significant fraction of the largest plumes can be detected automatically with negligible false positives.  This represents a significant new high-water mark in plume detection accuracy. Second, we use it for retrospective analysis to find plumes that were missed by the existing human review process. We observe that at least 25\% of large plumes may have been passed over in the existing workflow due to confirmation bias and ambiguity in the visual cues used by human reviewers. Finally, we extend detection to three understudied trace gases: \nhthree, \notwo and the first observations of carbon monoxide (CO) plume in EMIT imagery.
\end{abstract}
\begin{document}

\flushbottom
\maketitle
\thispagestyle{empty}

\section{Introduction}

\begin{figure}[!t]
    \centering
    \includegraphics[trim=0 0 0 0,clip,width=1.0\textwidth]{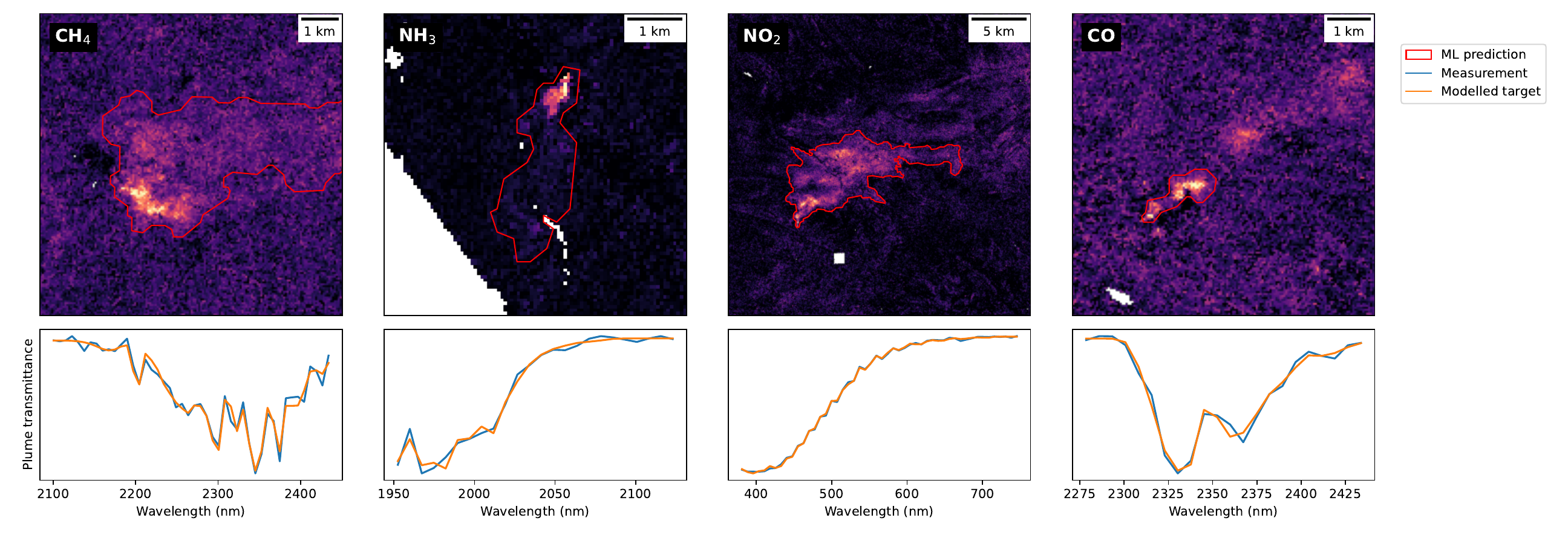}
    \caption{Examples of automatically detected trace gas point sources with their respective enhancement products and the machine learning model predictions outlined in red. Below we also show computed spectral fit plots with the modelled target signature of each gas species. Emission sources include a landfill in Chile (CH$_4$), a fertilizer plant in Nigeria (\nhthree), power plant in Iran (\notwo), and a metal and chemical production facility in China (CO).}
    \label{fig:example_trace_gas_plumes}
\vspace{-2mm}
\end{figure}

Orbital imaging spectroscopy with fine spatial sampling has emerged as an effective tool for finding trace gas point sources.  It supports rapid detection and mitigation of methane leaks occurring at malfunctioning oil and gas facilities, waste management landfill sites, and coal mines to remove a significant radiative forcing agent \cite{ruzicka2025operational, lee2023ipcc, kuylenstierna2021global_UN, national2019thriving_DecadalSurvey}.  Orbital monitoring can also help mitigate other gases like \notwo (Nitrogen dioxide), \nhthree (Ammonia) and CO (Carbon monoxide) which are important as air pollution and health risks \cite{taha2025comprehensive}.
Elevated levels of \notwo can lead to respiratory issues and exacerbate asthma \cite{world2006air}, while \nhthree can contribute to the formation of particulate matter, posing additional dangers to cardiovascular health \cite{committee2004air}; CO exposure can impair oxygen delivery in the body, leading to serious neurological and cardiovascular effects \cite{epa1991air_co}.
Examples of detections of these gases made automatically by our system are shown on Figure \ref{fig:example_trace_gas_plumes}.
Imaging spectrometers in the visible-shortwave infrared range from 380-2500 nm are well suited to the task, with sensitivity to localized enhancements and the spatial sampling to pinpoint their sources. These include narrow-swath instruments like Carbon Mapper Tanager-1 that perform targeted acquisitions at specific sites \cite{duren2025carbon_tanager}, wide swath instruments like \fixM{NASA's Earth Surface Mineral Dust Source Investigation (EMIT)\label{rev2_q5}}{R2-Q5} that cover large areas of the globe  \cite{thorpe_attribution_23}, and in the future, still more capable instruments that will cover the globe with rapid repeat cadence. NASA's \fix{Explorer for Artemis Geology Lunar and Earth (EAGLE-VSWIR)} \cite{cawse2021nasa_SBG} and ESA's \fix{Copernicus Hyperspectral Imaging Mission for the Environment (CHIME)} \cite{nieke2023copernicus_CHIME} missions will provide sub-monthly repeat intervals, expanding existing data volumes by over an order of magnitude and revealing thousands of hitherto-unknown sources.

While these instruments portend important benefits for mitigating pollution, their huge data volumes create new challenges. Most current workflows rely on manual review to spot and delineate gas plumes in the retrieved concentration field. Analysts must also discriminate plumes from false positives that arise due to structured background clutter or instrument noise. With the next generation of instruments acquiring over a million spectra per second, these manual reviews will be infeasibly laborious and expensive. Moreover, manual vetting is subject to a degree of bias, as analysts with limited time focus more on areas where sources have been spotted before. When assessing a candidate detection, analysts may be more likely to confirm an ambiguous plume that is associated with visually-suggestive ground infrastructure. The extent and impact of these biases has not yet been quantified. Confirmation bias extends to instrument tasking, where --- for narrow swath instruments --- acquisitions are scheduled where plumes are expected to be found. Thus, maximizing mitigation potential may also lead to under-reporting in areas that have been less historically active. Finally, manual review introduces a slow human bottleneck into processing, lengthening the latency between acquisition and actionable information.

Machine learning (ML) systems have been developed to assist the manual review process, making analysts more efficient and potentially also more objective. For example, a system for detection of trace gases and monitoring of methane point sources was operationally deployed at the United Nations MARS group by \cite{ruzicka2025operational}, where it proposes candidate plumes for human analysts to confirm. However these systems still produce a large number of false positives, since structured background clutter can look similar to small plumes with concentrations near the detection limit. Thus, this system still depends on human analysts to review and manually filter the automated results. Despite many proof of concept studies on small benchmark datasets \cite{jongaramrungruang2022methanet, groshenry2022detecting, joyce2023using, kumar2023methanemapper, ruzicka2023starcop, bue2025towards}, fully automated detection of even large methane leak events remains elusive. No system has yet achieved operationally-useful accuracy on a mission-scale dataset without significant human intervention. In large part, this is because the tolerance for false positives is historically low.  With hundreds of thousands of scenes being processed, and true leaks being globally rare, a machine learning algorithm with a 1\% false positive rate might still provide more false positives than true plumes. Consequently, the latency of the detection still largely depends on manual review, which can linger weeks or months behind the acquisition.

This paper presents a new approach to plume detection which is fully automated and global in scope, by focusing on delivering as many large plumes as possible while maintaining low false positive rates. We describe its deployment for NASA's EMIT instrument on the International Space Station. 
Our approach uses machine learning models to detect gas plumes in enhancement images, but then scores the resulting candidates using physics-based plume transmittance models. 
\fixM{Each of these steps have been used in isolation in the search for trace gas events, however for practical use we find that the combination is critical for fully automated use. The matched filter alone produces too many false alarms as shown in \cite{ruzicka2023starcop, ruzicka2025operational}, avoiding the dependency on the matched filter product as shown in end-to-end machine learning models of \cite{HyperspectralViTs} sacrifices the generalisation capabilities of the models and thirdly the spectral fitting approach of \cite{xiang2025identification} requires plume delineations in order to be automated.\label{rev2_q2}}{R2-Q2}
This combination sacrifices the identification of every plume in order to provide sufficiently robust identifications for use in operations without the need for human intervention, a high water mark for accuracy and our standard for considering a system to be fully automated.  The system is demonstrated to be capable of spotting large point sources in the operational context, while also revealing previously missed plumes from the archive. 
This makes the automated system complementary, rather than strictly subordinate, to the human review process. 

This method is deployed in two distinct modes of operation. First, we report its use as a fully-autonomous plume detection engine. We deploy it on a daily basis to analyze EMIT data immediately after downlink, producing a digest of recently-imaged point sources for fast response. This reduces the time between event observation and discovery from up to few months to approximately one day.
The deployment automatically discovered \dailyFoundEvents large methane point source events over its first \monthsRunningDaily months of operation, allowing detections to be made in a timeframe that was useful for notification as described in \cite{ruzicka2025operational}. Second, we deploy the system in collaboration with human reviewers to prioritize candidates for manual assessment. This improves the efficiency of human reviewers, significantly reducing the number of human hours required to find plumes. It also combats bias by prioritizing scenes using objective characteristics of the plume, and by informing analyst review with spectral fit metrics beyond purely morphological cues. 
Combining physics and machine learning far outperformed a pure machine learning method, discovering \archiveCorrectPrevMissed previously omitted methane point source events in 2 years of historical data from the EMIT archive. This suggests that at least 25\% of high confidence plumes may have been passed over in the purely manual workflow due to built-in conservatism in the review process.

In addition to the previously explored detection of methane, we adapt our system for the detection of other trace gases such as \nhthree, \notwo and CO, following the same daily monitoring paradigm.
Prior literature has demonstrated the capability of \nhthree and \notwo point source detection from imaging spectroscopy data in a limited number of instances - about 3 locations for \nhthree using Tanager-1 in \cite{balasus2026mapping} and 2 locations for \notwo using EnMAP in \cite{borger2025no2co2}. Our work deploys the hybrid machine learning and physics-based approach developed for methane retrievals in order to significantly expand the number and range of these detections.
We also report the first ever detection of CO plumes in EMIT data and discover in total \coEvents point source CO events globally.
Our analysis discovers that these gases are more difficult to detect than CH$_4$; the strengths and shapes of their absorption features are not as distinctive, meaning that the same system does not achieve full automation for these targets as it does for CH$_4$. Automation may yet be achieved by future algorithms and/or sensors with greater radiometric sensitivity or finer spectral sampling. 
Nevertheless, many more plumes are discovered than were previously known, further demonstrating the complementary nature of morphological and spectral cues for plume detection and supporting this unique union of machine learning and physics-based methods.

\section{Methods}
\label{methods}

\begin{figure}[!t]
    \centering
      \includegraphics[trim=0 2.3cm 0 3.4cm,clip,width=0.85\textwidth]{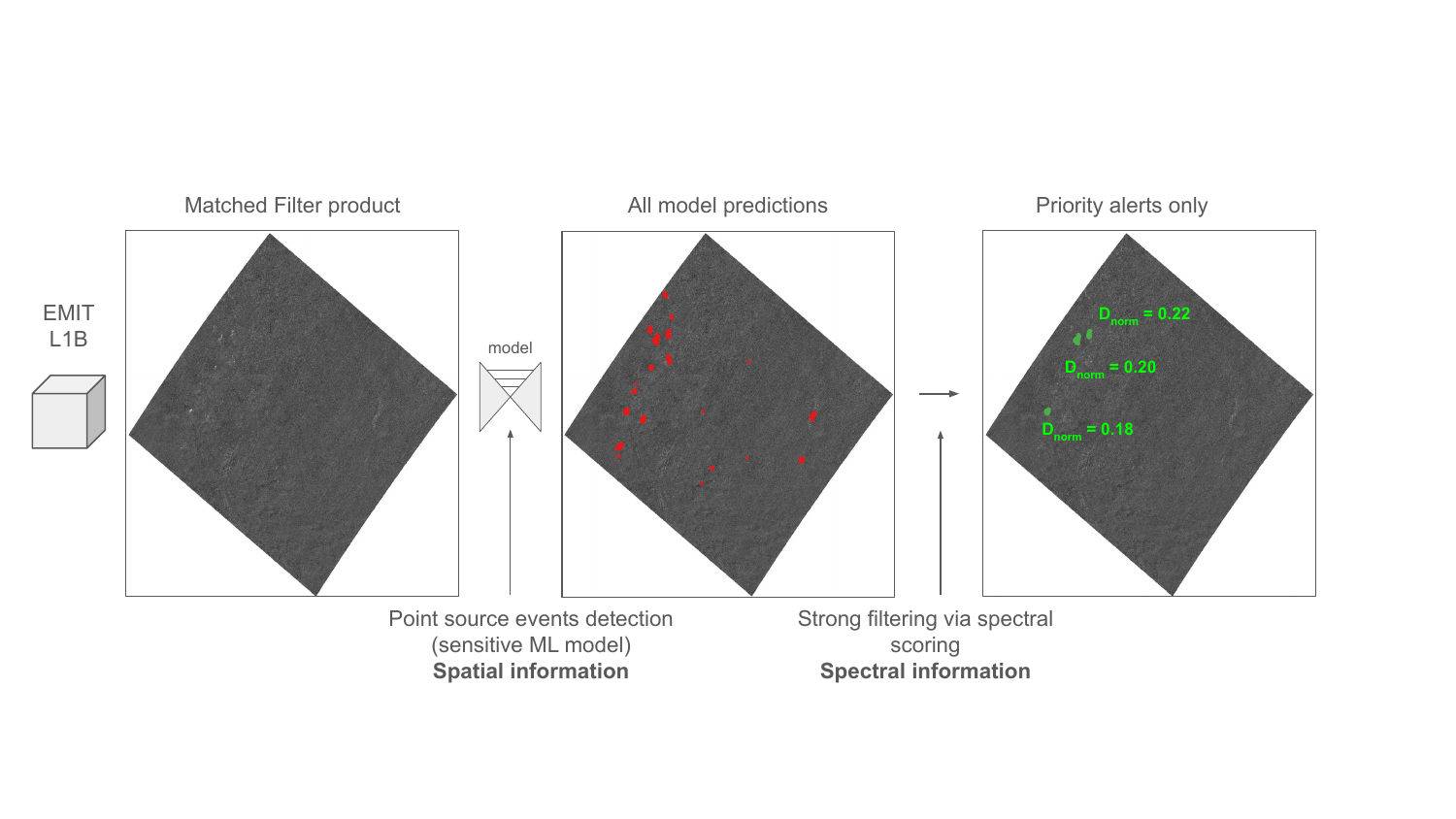}
    \caption{Method overview. Our proposed processing pipeline starts from radiance data, computes matched filter enhancement products and uses machine learning models to generate initial prediction proposals. Then we use spectral fitting metrics to filter our low fit score examples, leaving only high confidence predictions. While this example shows a scene with methane point sources, the same general method can be used with a number of other trace gas targets (such as \nhthree, \notwo and CO).}
    \label{fig:method_overview}
\vspace{-2mm}
\end{figure}

Figure \ref{fig:method_overview} introduces the general overview of our method. 
We start with spectral \fix{L1B} radiance data from the NASA's orbital imaging spectrometer EMIT. \fixM{While some variants of machine learning models can operate directly on radiance data \cite{HyperspectralViTs}, we instead\label{rev2_q1}}{R2-Q1} compute matched filter products for each target trace gas. Then, we use state-of-the-art machine learning models from \cite{ruzicka2025operational} and \fix{adapt them to our matched filter products} to generate candidate plume outlines. 
We show that while the models were trained for methane, they generalise to other trace gases. These initial predictions contain many false alarms alongside true point source detections. Finally, we use the background ratio method of \cite{xiang2025identification} to determine the transmittance of the plume and compare this to the known signature of methane absorption. This physics-based, spectroscopic step effectively rejects false alarms while preserving high confidence predictions.

\subsection{Matched filter feature extraction}
\label{method_mf}

The matched filter (MF) is a a common enhancement product used to find trace gases and other subpixel targets in imaging spectroscopy data \cite{thompson2015real}. In this case, it is used to estimate a target gas (e.g. methane) concentration above background levels in the scene.  The MF operates over a specific user-selected spectral range. Each trace gas target has specific recognizable absorption features; for example, methane has a weak absorption from \fixM{1650-1700 nm\label{rev2_q3}}{R2-Q3} and a stronger one from \fix{2100-2440 nm}. Work such as \cite{roger2023wide_MF, bue2025towards} show that variants using extended spectral ranges outperform the default MFs that work only on the narrow range of each absorption feature. We use the wide-window matched filter (WMF) of \cite{roger2023wide_MF} for methane and the extended column-wise matched filter (CMF) for other trace gases.
The work of \cite{ruzicka2025operational} quantitatively compares MF variants on large scale datasets, further highlighting that implementations using wide spectral range outperform narrow and other iterative approaches. Practically, these matched filter products are also used by analysts to manually search for methane leak events - the WMF product at UN MARS unit and the CMF product in the JPL analysts team.

The MF routine takes a subset of the full spectral radiance data products (for EMIT this is 285 bands) and produces a single band output representing the gas enhancement above background levels as a mixing ratio length.  The Visible-ShortWave InfraRed (VSWIR) range contains vibrational overtone absorptions of many trace gases.
In this paper we focus on \chfour, \nhthree, \notwo and CO.

The MF is implemented via a noise-whitened dot product between a target absorption spectrum and each mean-subtracted test spectrum in the image.   We generate target spectra for \chfour, \nhthree, \notwo and CO using MODTRAN \cite{berk1987modtran} to simulate spectral radiance with and without the gas as was done for the data product of \cite{emitl2b_atbd}. These enhancements relate to emission rates via an unknown windspeed-dependent scale factor.  Here, we will focus simply on plume detection using the relative enhancement above the background, and save emissions estimation for future work. 
\fixM{However, in case of \chfour, we use the IME method \cite{frankenberg2016airborne, varon2018quantifying_IME} with the ERA5 10m wind speed data \cite{hersbach2020era5} for quantification.\label{rev2_q6a}}{R2-Q6a}

\subsection{Semantic segmentation with machine learning models}
\label{method_semseg}

For the initial automatic plume detection, we leverage state-of-the-art semantic segmentation models proposed by \cite{ruzicka2025operational}. The machine learning model uses a U-Net architecture \cite{ronneberger2015u_unet} with the MobileNetV3 encoder \cite{howard2019mobilenetv3}. The models have been trained on RGB bands extracted from the radiance cubes and the WMF product computed for methane using the MARS EMIT dataset published in \cite{ruzicka2025operational}. We use an ensemble of 5 trained models which significantly reduces the false alarm rate.

These models enable zero-shot generalisation to other trace gases. We omit the dependence on the RGB bands (as those might include information specific to individual gas only), and keep just the matched filter as the single input band.
Matched filter works as a sort of general interface between different gases. \fixM{For example, a model trained on radiances directly (e.g. \cite{HyperspectralViTs}) would learn to identify absorption features specific for the trained gas and therefore wouldn't generalise to other gases. Meanwhile the matched filter product looks generally \textit{similar} between different target gases, and this is why the models generalise.\label{rev2_q4}}{R2-Q4}
\fixS{between complex hyperspectral data and our plume shape detecting machine learning models.}
Examples of MF products on Figure \ref{fig:example_trace_gas_plumes} motivate us, that a general plume shape detector would be able to generalise between different trace gases.

For the detection of \nhthree, \notwo and CO, we train machine learning models on the task of \chfour semantic segmentation using the MARS-EMIT dataset \cite{ruzicka2025operational}, following the same used hyperparameters. However, instead of using the packaged WMF product, we re-compute the CMF product for each scene. 
We normalise the input values into range expected by the machine learning models.
This choice is made to match the variant of the matched filter used for other trace gases. 
As such, the domain shift distance between \chfour and other gas species is lower than if we used two different matched filters (for example WMF to CMF).
We train a total of 4 new machine learning models and use them in an ensemble.

Regardless of the gas target (so either \chfour or the other trace gases), we follow the same post-processing setup. We threshold final model ensemble predictions using the default value of 0.5, and identify plumes using connected components. 
We do not post-process these instances further (e.g. we do not merge nearby predictions), saving as many possible candidates as possible for the physics-based false rejection step.  
We however found that size of the predictions can also be used to filter out predictions triggered on small artifacts in the MF products.

For trace gases other than methane, we apply the models directly without retraining. Similar zero-shot generalization between sensors was showcased in prior work \cite{ruzicka2025operational, ruzicka2025intelligent, ruzicka2023starcop}. Here, we explore generalization between different gas species, applying methane plume models to other trace gases such as \nhthree,  \notwo and CO. The current bottleneck to tuning models to other gases is the limited knowledge of these point sources globally and consequent paucity of training data. This shortage makes zero-shot generalization critical to build high-accuracy models for these understudied point sources.

\subsection{Physics informed spectral scoring}
\label{method_spectral_scoring}

\begin{figure}[!h]
    \centering
    \includegraphics[width=0.7\linewidth]{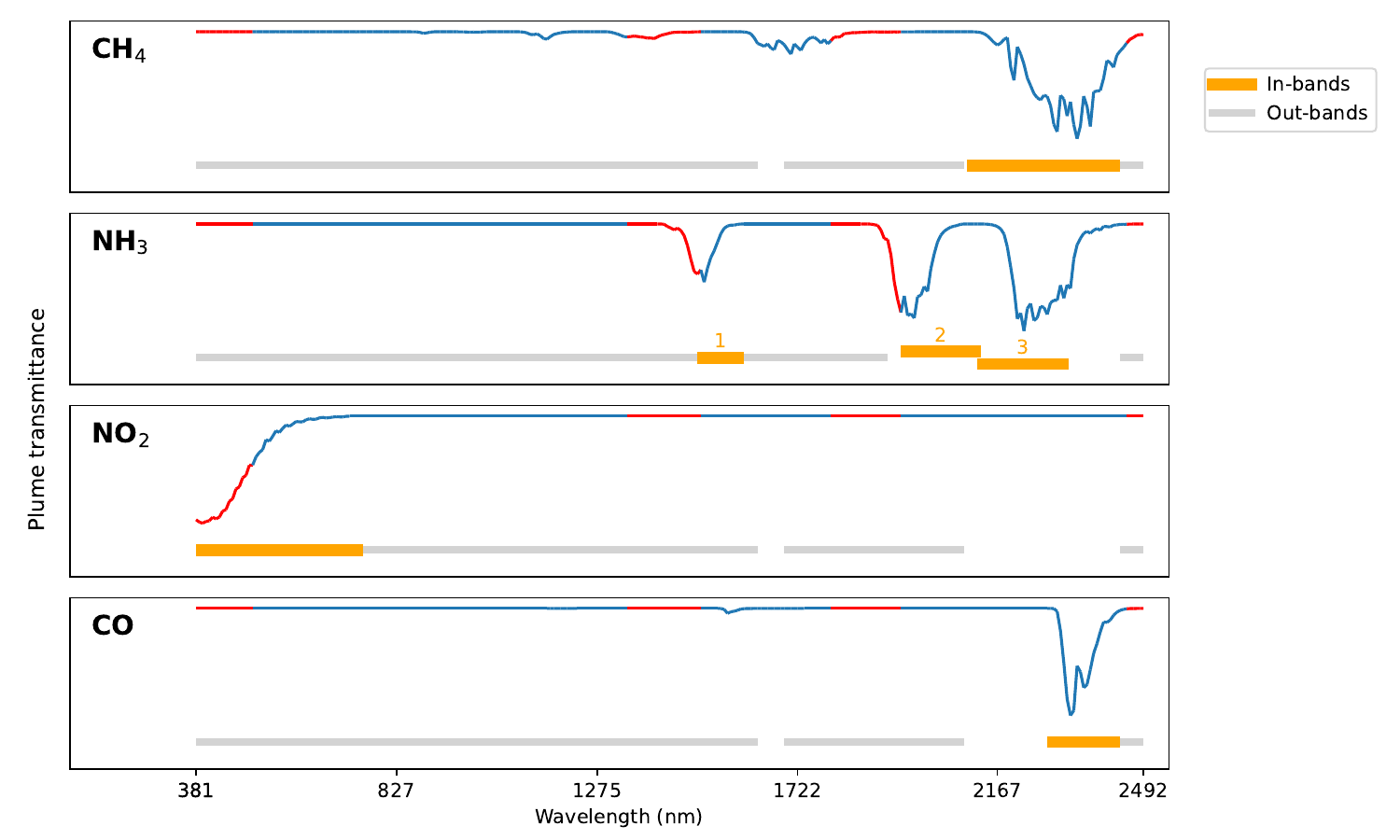}
    \caption{Spectral fit scoring for different trace gas species. In orange, we show band regions used for comparison with the known target signal (for \nhthree in three separate sections) and in gray the band regions used as background of the signals which are used to match similar hyperspectral pixels outside the predicted vector outline. Red colour shows the regions of the data which is not used when computing the MF enhancement product (however as can be seen from orange regions they can still be useful for spectral fit computation).}
    \label{fig:gas_species_targets_ranges}
\vspace{-2mm}
\end{figure}

The work of \cite{xiang2025identification} validated manual methane point source detections by measuring the aggregate plume transmittance. Here we extend the approach to score plume outlines produced by the previous machine learning step. We extend it to trace gases other than methane as illustrated on Figure \ref{fig:gas_species_targets_ranges}. We briefly summarize the steps taken during the scoring. More details can be found in the supplementary materials and in the original paper \cite{xiang2025identification}.

Crucially, plume scoring depends on data products from all previous steps. It requires polygonal outlines of suspected methane plumes (predicted automatically or manually drawn by analysts), the matched filter product for selecting points inside and outside of each plume, and finally the radiances for accessing the spectral information at each pixel.  In contrast to the plume segmentation step, which applies pattern recognition in the spatial domain to recognize the morphology of plume enhancements, this stage uses physics-informed knowledge of the absorption signatures of the target gases in the spectral domain. 

For each candidate plume outline, either a labelled event from a manually-drawn catalogue, or from the predictions of machine learning models, we measure the transmittance using a ratio of the in-plume pixels to the local background outside the plume. We select 40 points inside, extend them into their local neighborhoods and match each point with a spectrally similar point outside of the plume boundary. We use the resulting pairs to compute an in-plume to out-of-plume ratio that represents the plume transmittance along the measured optical path. As shown in Figure \ref{fig:gas_species_targets_ranges}, different trace gases use different configurations of the absorbing and non-absorbing intervals. In the case of \nhthree, which has multiple diagnostic features in the VSWIR range, we consider three separate regions for the absorbing intervals.

Following \cite{xiang2025identification}, the model fit gives us two measures of the plume strength: the aggregate plume mixing ratio length and a spectral fit score that indicates how well the model matches the data.  Our spectral fit score is the mean absolute error, i.e. the L1 norm, which is more resilient to non-Gaussian outliers than a classic squared error.  We will refer to this score as \dnorm. A plume is likely to have low \dnorm indicating a good model fit with a large mixing ratio length indicating a true plume enhancement. Our work thus expands on that of \cite{xiang2025identification} with experiments on a large labelled dataset of methane sources, assessing \dnorm scores for actual methane point sources and non-sources caused by background clutter. Additionally, we use this method in conjunction with machine learning models and extend it to other trace gases.

\subsection{Data}
\label{method_data}

In this paper, we use data from NASA’s EMIT imaging spectrometer, but our methods could be extended to other sensors as well.  The EMIT instrument began its mission observing the geology of arid mineral dust source regions from the International Space Station.  After completion of this prime mission in 2024, it began observing the entire land area of the globe that is reachable by the International Space Station.  Its combination of high uniformity and sensitivity provides good sensitivity for detecting trace gases \cite{thompson2024orbit}. 

To assess the method, we use the MARS-Hyperspectral v2025 dataset published in \cite{ruzicka2025operational}. We use this benchmark dataset for two purposes.  First, it contains a pre-validated list of days with large known methane sources used to construct a validation dataset for our method. We select 21 largest emissions in the dataset and query all EMIT scenes captured in the -12 and +12 hours range (selecting a full day worth of data). We also select 22 days without any a priori known events and use these to balance the set with negative (no-plume) scenes. Note, that while we used the largest methane sources to select these days, it also contains many smaller methane leak events, for which we keep the labels. This set contains 3175 EMIT scenes.

In addition, we construct a second set with all 7.2k EMIT scenes from the MARS-Hyperspectral v2025 dataset to guide our search for other trace gases.  These data are sampled globally and over the whole period of EMIT data archive (ending with events until the end of 2025). These scenes are collocated with methane emitting infrastructure, but they are also balanced to contain negative scenes with no methane plume present. Importantly, we don’t use this second set for the \chfour experiments, we use it to compute products for other trace gases, following the intuition that centres of population and industry that are emitting methane can also emit other gas species. Other heuristics can be used in the future to further expand the search, but prior to this work, only few emitters were known for these other species \cite{balasus2026mapping, borger2025no2co2, kuai2019aerial_nh3}.

\section{Results}

We divide the results into four distinct applications of the proposed method. First, in Section \ref{res_method_verif}, we assess the performance of our combination of ML-based detection with physics-based vetting using a large, representative dataset of methane leak events. Second, in Section \ref{res_daily}, we present the outcomes from an ongoing deployment of the Daily Methane Digest system, which uses our method operationally to detect large methane sources with improved time-to-information in contrast to prior manual methods. Third, in Section \ref{res_archive}, we demonstrate the approach for the task of archival completion of the methane events catalogue. Finally, in Section \ref{res_other_gases}, we explore the use of our system for the detection of other trace gases, demonstrating the first ML-assisted detection of \nhthree, \notwo and CO point sources from the EMIT instrument. 

\subsection{Method validation}
\label{res_method_verif}

\begin{figure}[t]
  \centering
  \begin{subfigure}[b]{0.45\textwidth}
    \centering
    \includegraphics[width=\textwidth]{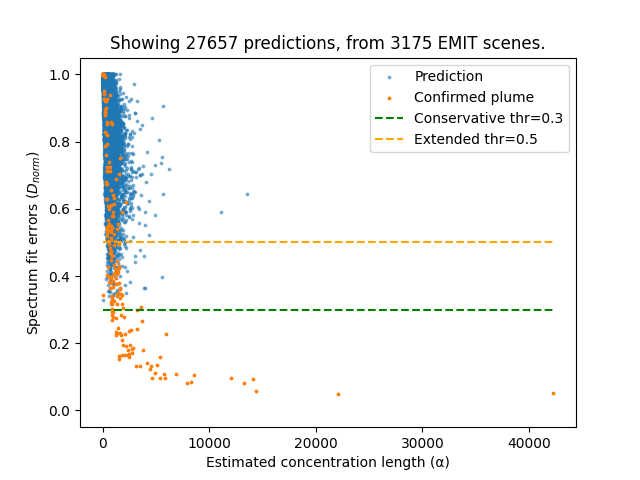}
    \caption{Prediction scores of \dnorm and $\alpha$.}
    \label{fig:res_method_validation_left}
  \end{subfigure}
  \begin{subfigure}[b]{0.45\textwidth}
    \centering
    \includegraphics[width=\textwidth]{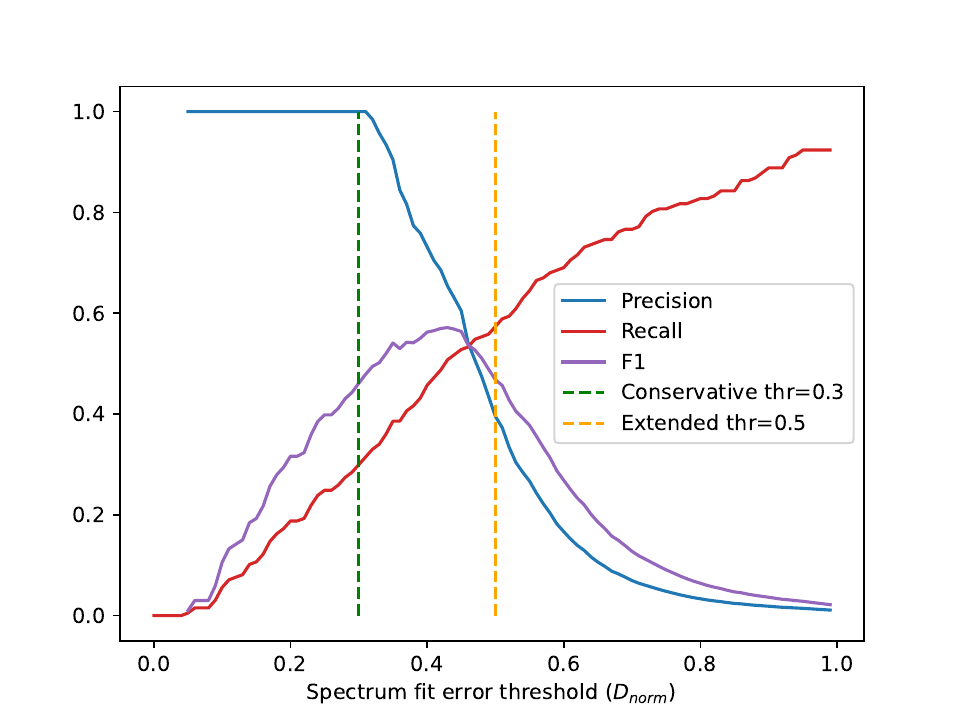}
    \caption{Statistics depending on the threshold on \dnorm.}
    \label{fig:res_method_validation_right}
  \end{subfigure}
  \caption{Exploring the distribution of scores assigned to all predictions in the validation set of selected days (total of 3175 EMIT scenes). Any prediction which overlaps with a known ground truth label is marked as confirmed and shown in orange (on the left). On the right, we show the Precision, Recall and F1 classification statistics depending on the used threshold on \dnorm. We also highlight the thresholds of 0.3 and 0.5, as we use these in other experiments. Note that we include only predictions larger than 25 pixels.}
  \label{fig:res_method_validation}
\end{figure}

To validate our proposed methodology, we created a dataset of 3175 EMIT scenes sampled from selected days, as described in Section \ref{method_data}.
We highlight that the size of this method validation dataset is order of magnitude larger than the previously used validation datasets in \cite{ruzicka2025operational} (which considered total of 200 scenes).

For each scene we compute the wide-window matched filter (WMF) of \cite{roger2023wide_MF} and we extract RGB bands from the EMIT radiances. We use the machine learning model ensemble released in \cite{ruzicka2025operational} to create candidate plume outlines. These predictions include a model confidence score calculated in the following manner: (1) we apply a threshold to continuous model ensemble predictions (2D product with data range of 0-1) at 0.5; (2) we connect contiguous detected pixels into predicted plumes; and (3) we reduce the threshold as needed so that each selection contains 5 pixels. The final threshold which selects at least 5 pixels is then used as the model confidence score.
The used values can be adjusted to change sensitivity. For example, the initial threshold can be selected higher than 0.5 to filter only detections where the model is more confident.  Similarly, the number of pixels can be adjusted according to the desired plume size. We note, however, that these settings worked well for our experiments.

Next we calculate the spectral metric scores for each predicted plume candidate. In summary, we compute the aggregate plume transmittance by dividing pixels inside and outside of the delineated plume vector, and then calculate the spectrum fit error between this ratio and the known gas absorption.  This score is called \dnorm. We also get the estimated concentration length measured in ppm m, which we call $\alpha$.  Notably, this step depends on the availability of a plume boundary identified through other means, so it can only be automated with an ML-based or similar approach for producing the initial candidate. 

Figure \ref{fig:res_method_validation_left} shows the distribution of these scores for the whole method validation set. Predictions which overlap with known ground truth labels are marked as confirmed. We see that there is some overlap between distribution of true methane sources and non-plume predictions, but there are also regions of the parameter space that are purely one or the other. Importantly, using the relatively conservative threshold of \dnorm= 0.3, all predictions with lower scores are true point sources. We denote these detections as highly certain, and use it in other experiments to automatically trigger alerts. A second interesting region of this parameter space is \dnorm $\in (0.3; 0.5\textgreater$. As seen on Figure \ref{fig:res_method_validation_right}, while the methane point sources are intermixed with non-plumes, one can select this higher threshold to capture a larger proportion of potential point sources.  This allows users to adjust the system according to the specific operational needs and available analyst time. 
Additionally, Figure \ref{fig:res_method_validation_right} explores the amount of missed manually labelled methane point source events according to the selected threshold value. We note here, that the used labels include many small and hard to detect plumes - this was a consciously made choice to mimic the difficult challenge of real world deployment.

\begin{figure}[t]
  \centering
  \includegraphics[trim=0.3cm 0 0.5cm 0,clip,width=0.32\textwidth]{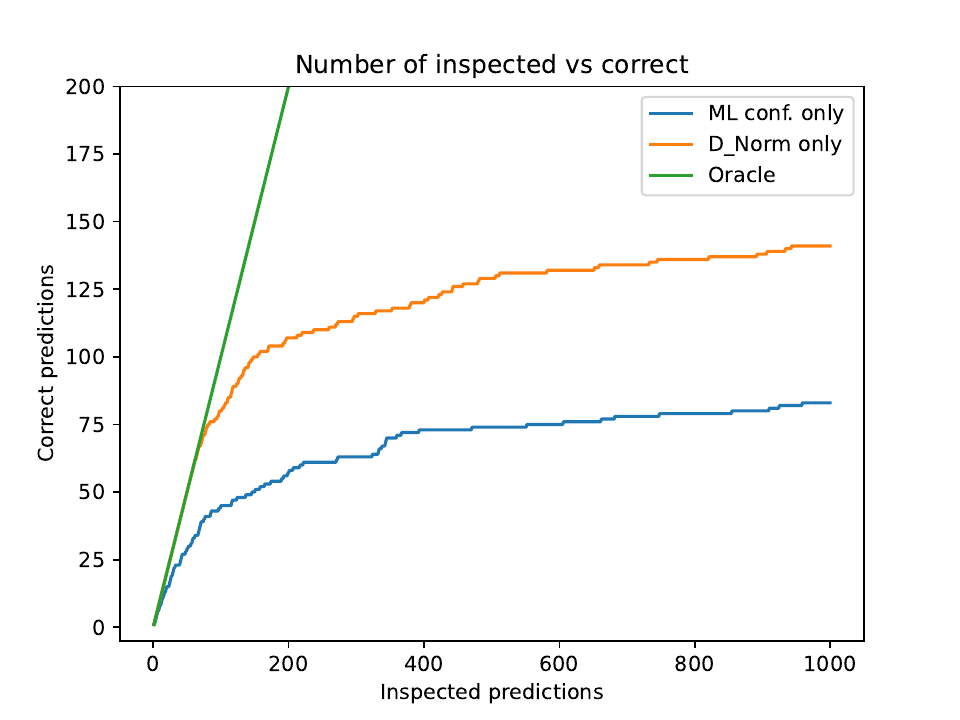}
  \hfill
  \includegraphics[trim=0.3cm 0 0.5cm 0,clip,width=0.32\textwidth]{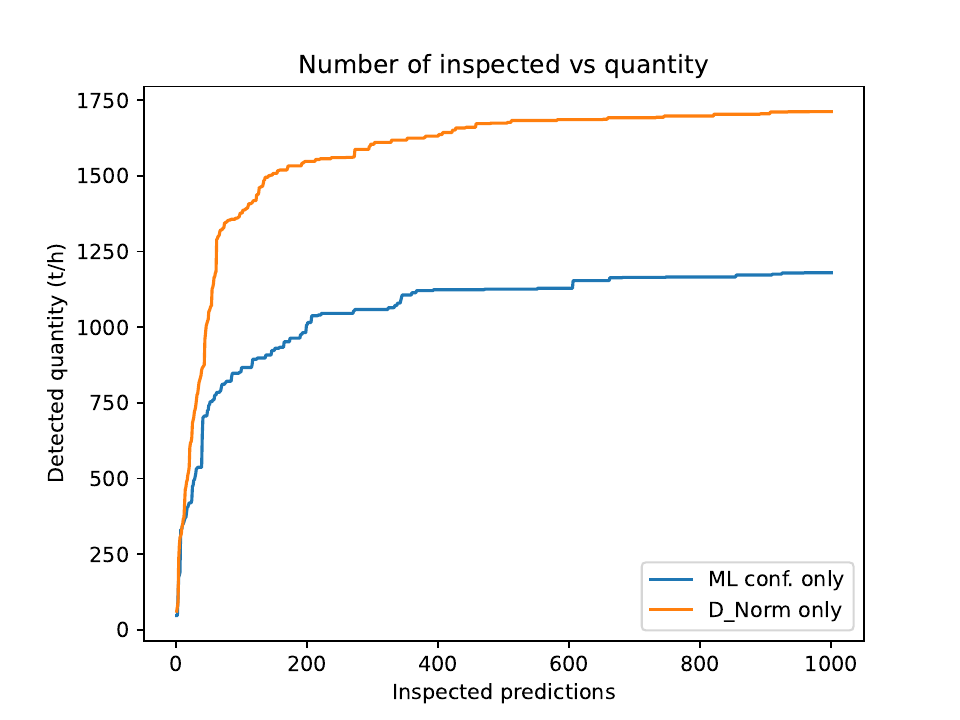}
  \hfill
  \includegraphics[trim=0.3cm 0 0.5cm 0,clip,width=0.32\textwidth]{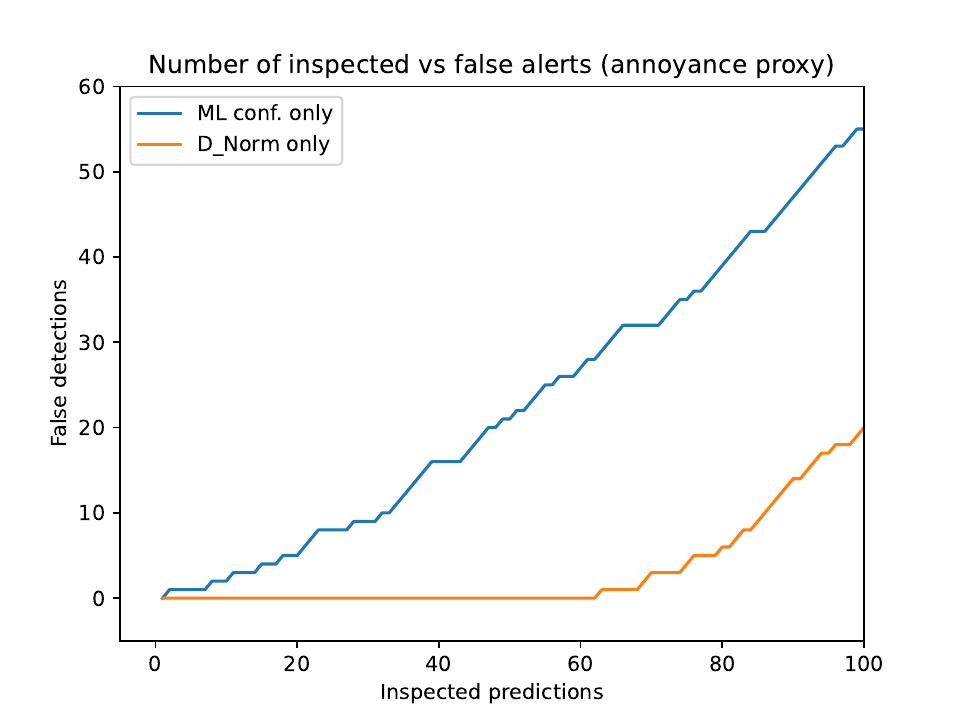}
  \caption{Visualisation of the amount and quantification of detected methane sources under a specific number of inspected predictions (left and middle). We compare machine learning only approaches (using the confidence score of the model) with our proposed system which combines machine learning predictions with spectral fit scores. On the right, we show a analysts annoyance proxy - number of false alarms that are present in the inspected predictions. Note that we include only predictions larger than 25 pixels.}
\label{fig:res_prioritization}
\end{figure}

Figure \ref{fig:res_prioritization} shows a more continuous approach to plume scoring. In an operational setting, we can score each prediction with these metrics (model confidence, \dnorm and $\alpha$) to prioritize candidates for manual review. The order of inspections is given by the prioritization function.  This approach can guide analysts to more and larger point sources than the machine learning model confidence scores alone. To estimate emission rates \fixM{of \chfour\label{rev2_q6b}}{R2-Q6b} we used the IME method \cite{frankenberg2016airborne, varon2018quantifying_IME} with the ERA5 10m wind speed data \cite{hersbach2020era5}. The spectrum fit scores are highly impactful when focusing on the most confident predictions. 
In this regime, the spectral fit scores closely follow the performance of a perfect ``oracle'' with each inspection corresponding to a detected event and negligible false positives.
This property is important for automated daily alerts, where the annoyance cost of false positives is high. A daily alert that relied only on machine learning model confidences would almost immediately produce high-scored but incorrect predictions.  This would quickly damage the credibility of the automated alert system.  
We observe that for 1000 inspected events, spectral prioritization outperforms the ML confidence score by 71\% in terms of the number of detected events and by 46\% in terms of the total plume mass.

Finally, we highlight that while the spectral fit scores significantly outperform pure-ML model confidence scores, the ML ensemble is needed in the first place to create the initial set of potential plumes. As such this system presents a combination of ML models with spectrally informed metrics. Both components bring unique and mutually complementary information - the ML model ensemble considers the morphology and the shape of the plume alongside with the values of the enhancement products, while the spectral fit score considers the spectral information present in the full imaging spectroscopy data.

\subsection{Daily deployment}
\label{res_daily}

Following on the positive results of the performance study, we set up a fully automated ``Daily Methane Digest'' system to process all EMIT scenes collected each day. Here we present the results of this system since its commissioning on 1 January 2026. The system automatically sends predictions with \dnorm scores under 0.3 to a team of analysts to be considered for the mission's growing catalogue of methane leak events \cite{emit_plumes_v002_data_product}.

The current manual methane point source detection practice is described in the Algorithm Theoretical Basis Document (ATBD) \cite{emit_plumes_v002_atbd}, and performed within an internal Multi Mission Geospatial Information System (MMGIS) portal.  First, an analyst called Reviewer 1 manually searches through the methane enhancement products computed for all EMIT scenes captured each day. They look for and delineate any point source plumes. Events labelled by Reviewer 1 are then validated in follow-up review rounds. Reviewer 2 considers temporal information and scene background statistics to either confirm or reject the candidate plumes. Finally, Reviewer 3 makes a comprehensive final analysis to validate a plume for publication.
This process specifically prioritizes minimizing false positives, and is effective but time consuming. It requires significant time investment from trained human analysts, without which it can be subject to higher latencies. 
Moreover, the intentional conservatism of requiring agreement across three reviewers means that small events may be excluded. Our system effectively serves as a fully automated Reviewer 0, scanning all scenes captured for the selected day and immediately alerting the human participants of any detections. As such, the analyst (in the role of Reviewer 1) is directly shown the events of interest, and can allocate their review time more efficiently. Furthermore, the spectral fit statistics used for scoring each prediction can be used by Reviewers 2 and 3 to disambiguate uncertain cases. 

Having a fully automated Reviewer 0 allows for a  significant reduction in the latency of detections, i.e. the time to information from an orbital sensor observing the event to the analyst knowing about it. In some cases, the previous system's time delay before an analyst saw the EMIT scenes was more than a month. In contrast, our system allows for reliable near same-day detection of very large methane leak events. This is especially important in cases of large accidental releases, where early detection can lead to notification of the operators on the ground and remediation of the underlying source \cite{ruzicka2025operational}.  EMIT was not designed as a low-latency system, and the downlink, processing, and ingestion into NASA's archives have physically-induced latencies that are unavoidable, which can vary from twelve hours to a week or more.

\begin{figure}[!t]
    \centering
    \includegraphics[width=0.9\linewidth]{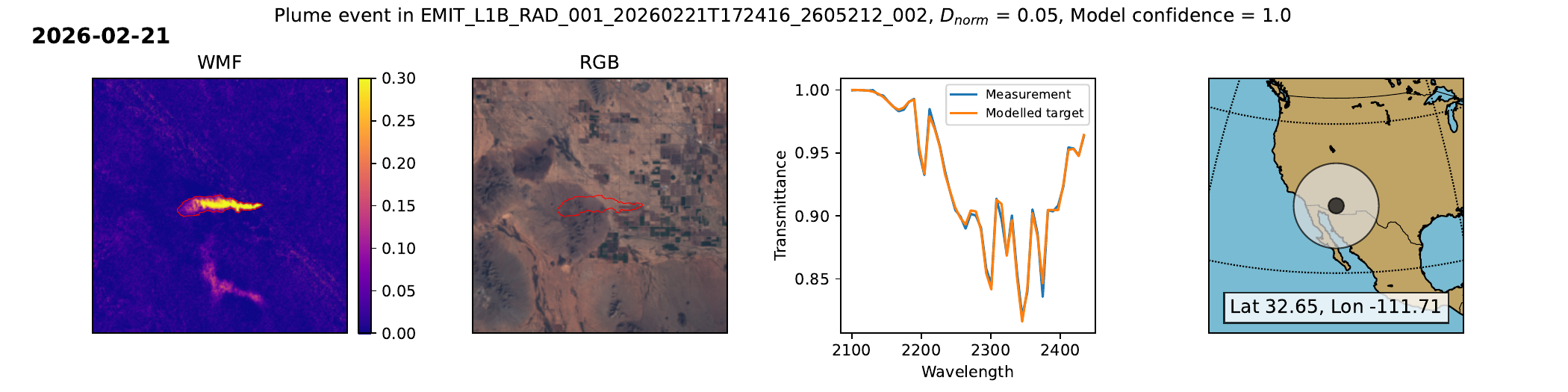}
    \caption{Generated report for a large methane leak from the 21st February 2026. From left to right we show the computed enhancement product, RGB bands from the EMIT observation, spectral fit plot and map with the location.}
    \label{fig:example_recent_detection}
\vspace{-2mm}
\end{figure}

Figure \ref{fig:example_recent_detection} shows one example of a very large recent methane leak from the February 21st detected using our system on February 24th. On the same day, the knowledge of this leak was shared with UN MARS group and the Carbon Mapper team \cite{duren2025carbon_tanager} that manually scheduled a follow-up acquisition with Tanager-1. Finally, on February 27th, the Tanager overflight was conducted, which showed no observable methane plume (scene name: \textit{tan20260227t191325c42s4001}). \fixS{indicating that the underlaying cause was likely fixed.}\fixM{\label{rev1_q1}}{R1-Q1}

The global deployment of the Daily Methane Digest enables the model to assess all locations in an unbiased way, in contrast to some EMIT data workflows that process only scenes overlapping previously known locations of emissions. 
Time consuming manual inspection of EMIT scenes can also lead to attention bias favoring locations or sectors that have produced sources in the past.  This is natural in presence of time pressure to process the data and the desire to detect as many events as possible in a limited time.  However, this can lead to under reporting of events in other areas.
While machine learning models can also introduce their own bias (for example if the models are being used in locations not included in their training datasets), we made a conscious effort to avoid the usual pitfalls by using global training data of \cite{ruzicka2025operational}, which spans a wide range of background substrates and sources.

\begin{table}[h]
\caption{Daily Methane Digest detections per month, reporting on events with \dnorm $< 0.3$.}
\label{tab:daily_detections_per_month}
\centering
\scalebox{1.0}{
\begin{tabular}{@{}lllll@{}}
\toprule
 & Processed scenes & Correct & Incorrect & Other \\ \midrule
January & 2240 & 7 & 1 & 0  \\
February & 4100 & 25 & 0 & 1 (potential plume tail) \\
March & 1636 & 11 & 0 & 1 (uncertain) \\ \bottomrule
\end{tabular}
}
\end{table}

Table \ref{tab:daily_detections_per_month} shows the performance of the Daily Methane Digest system used in the first few months of deployment by the JPL team at NASA. All correctly detected methane point sources are added into the review process described in \cite{emit_plumes_v002_atbd}. 
Predictions labelled by the analysts as incorrect or unlikely include potential plume flares (these are not added into this catalogue, although their detection can still be useful for mapping active O\&G infrastructure), instances without easily discernible infrastructure on the ground and a prediction over a potential methane plume tail (disperse plume over a larger area which is disjoint from it's origin).  
We note that EMIT is hosted on the ISS and such has an irregular pattern of observed areas. 
Therefore, the number of usable scenes and detectable methane leak events is expected to differ from month to month.

While the estimation of time needed to process one day's worth of EMIT data is difficult to quantify (numbers of scenes differ as well as the regions that are observed), our analysts estimated an average prediction with high confidence plumes to take about 26±1 minutes to be fully verified and drawn as a new plume outline (we note that this was in the case of scenes with complex background).
Additionally, it takes on average 10.25±0.75 minutes to reject a false alarm prediction, especially in ambiguous cases where the analyst needs to explore the underlying infrastructure in high resolution layers and check previous EMIT scenes for similar patterns. 
This timing already considers only the high confidence alerted events and as such doesn't include cases where the rejection is trivial upon inspection.

Searching for methane point sources in the whole archive of one day worth of EMIT data requires the analysts to pan and zoom to each suspected source in the enhancement product in addition to any later labelling efforts. 
Over the deployment of our system (\daysRunningDaily days at the time of this writing), we generate about \avgAlertsPerDay alerts per day. This should take about \avgMinsPerDay minutes on average daily to both confirm and draw vectors for true detections and reject incorrect predictions. In the future this may be further accelerated by using the plume outline predicted by the model instead of re-drawing it manually.
In contrast, in takes on average about 0.61 minutes per scene in each day for the analyst to conduct the fully manual search for events (note that this would still be followed by additional time needed for labelling and multi-analyst review). The \dailyTotalScenes scenes processed in the Daily processing pipeline would take about \dailyHoursExplore hours to explore, and more to label the detected events.  
All together, we estimate it would require at least 60-70 minutes of analyst time each day, and more for review layers, to produce results comparable to the automated system.  
For the forthcoming global EAGLE-VSWIR mission, with more than 35$\times$ the coverage of EMIT, this translates to at least 5 dedicated analysts working continuously on manual plume detection, an infeasible increase in science data staffing.  

\subsection{Archive completion}
\label{res_archive}

\begin{figure}[!t]
    \centering
    \includegraphics[trim=0.5cm 0 0.5cm 0,clip,width=0.7\textwidth]{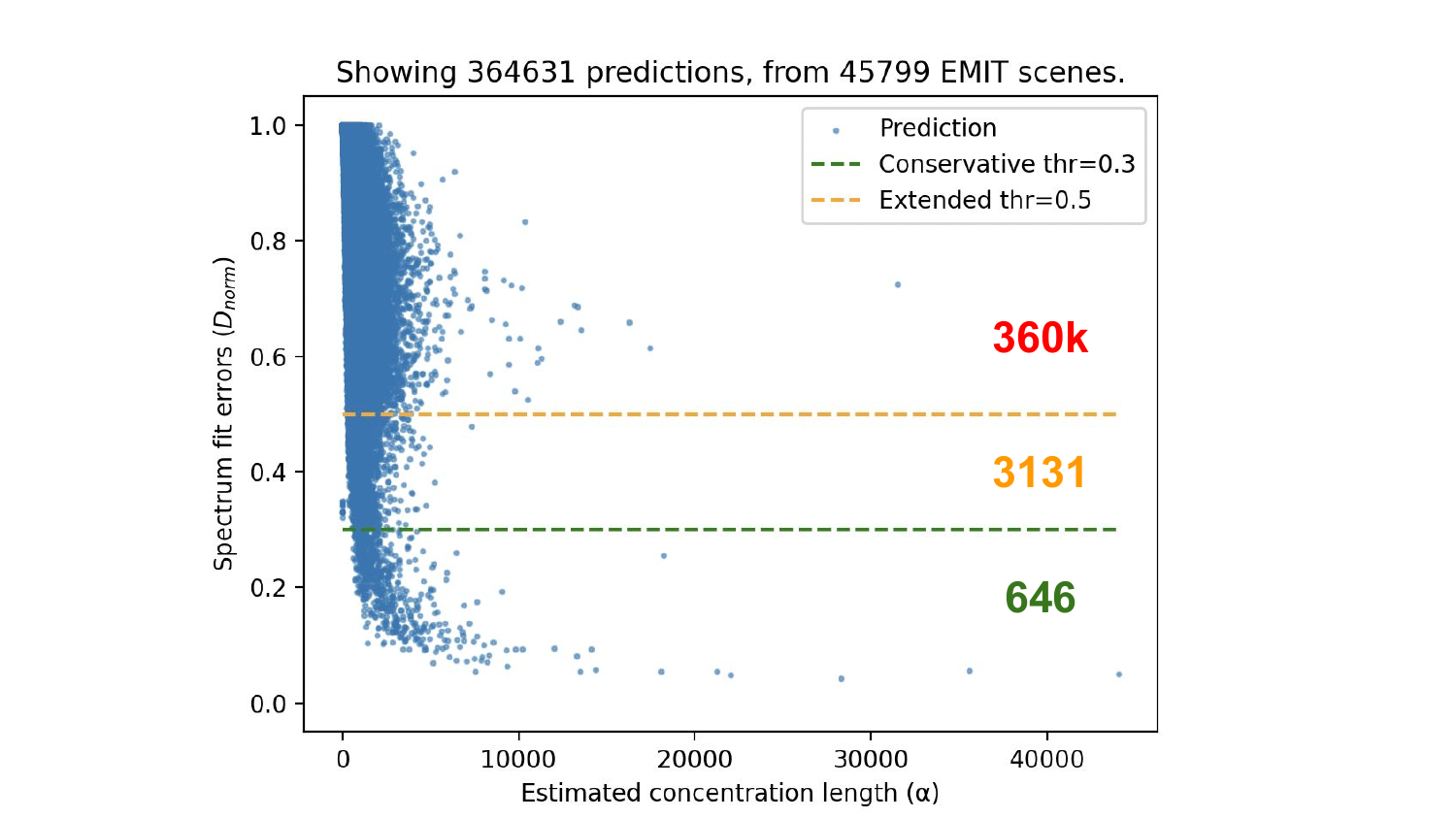}
    \caption{Scored predictions from 2 years of EMIT data, with 45.8k scenes processed. Note that we include only predictions larger than 25 pixels.
    }
    \label{fig:res_2years_archive}
\vspace{-2mm}
\end{figure}

As another use case of the system described in previous sections, we assess an entire archive of EMIT data for two selected years (2024 and 2025). For this experiment, we only consider scenes with cloud cover under 50\%. This process selects a total of 22986 scenes in 2024 and 23212 scenes in 2025; about a quarter of the total archive. We ensured that the ML models we use had been trained only on data from 2022 until 2023, preventing overlap with the training dataset of \cite{ruzicka2025operational}. The radiances for these scenes alone are estimated to be larger than 92Tb of data. We compute the wide-window matched filter (WMF) product of \cite{roger2023wide_MF} for each scene, use our machine learning model ensemble for inference, and compute spectral fit scores for each prediction. 
Using the conservative and extended thresholds of \dnorm from the initial results in Section \ref{res_method_verif}, we select three populations: high confidence (\textless0.3), low confidence ((0.3; 0.5\textgreater) and ignore category (anything above 0.5). Figure \ref{fig:res_2years_archive} shows the scores for the entire 2 years of data divided into these three categories.
We compare predictions to the labels in the NASA's catalogue of known methane leak events \cite{emit_plumes_v002_data_product} to explore if there were any events previously missed by the analysts. 

We manually inspect the high confidence set of \archiveHighPriorityAlarms high priority detections. 
Of these, 399 predictions overlap vectors previously proposed by the Reviewer 1 as methane point sources, while 247 were missing in these labels. First inspecting the 399 already known predictions, we find that 366 were released as confirmed plumes, 8 are in the backlog waiting to be reviewed and finally 25 have been labeled by a human reviewer but subsequently rejected in followup review.
Rejections are often not conclusive false positives, but are simply cases for which insufficient evidence was available to conclusively designate a point source.  For example, an apparent methane enhancement in a wilderness area with no visible infrastructure is likely to be rejected, even though new construction (predating the analysts' basemap) or pipeline leaks occasionally produce these signals. 

We next inspected the 247 cases which were detected by the automated system but missing in the catalog.  We manually determine that \archiveCorrectPrevMissed were correct methane leak plumes previously omitted in the manual labelling process. Of the remaining, 53 could not be confirmed or denied. This category includes flares, enhancement product artifacts, and other cases which were ambiguous even to the human inspection.  
We highlight that the computed spectral fit metrics could also enable confirming some of the events manually labelled as uncertain - seeing the spectral curve matching that of methane could be used as enough evidence in cases where no structure on the ground may be visible. 
Finally, 37 automated detections were rejected - there was not enough evidence to confirm these detections as plumes to be published as part of the \cite{emit_plumes_v002_data_product} product.

\begin{figure}[!t]
    \centering

    \includegraphics[trim=1.5cm 1.5cm 1.5cm 1.5cm,clip,width=0.9\linewidth]{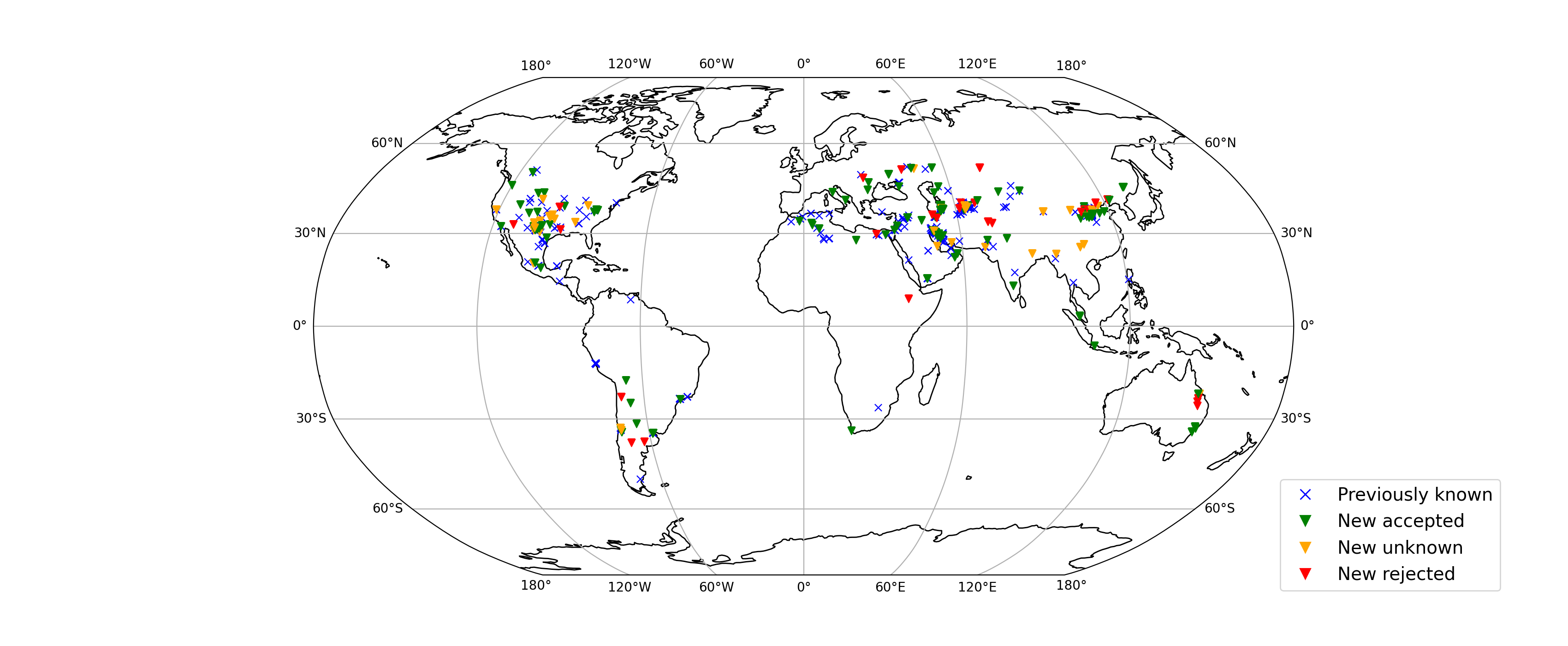}
    \caption{Map showing the distribution of \archiveHighPriorityAlarms high confidence alarms produced by our system between 2024 and 2025. In blue we colour the 399 predictions which overlap with previously labelled events.}
    \label{fig:map_new_events}
\vspace{-2mm}
\end{figure}

In Figure \ref{fig:map_new_events} we explore the locations of the previously missed large methane point sources in contrast with already known locations.
We observe new confirmed detections in several completely new regions and outline potential reasons for why these events might have been missed.
Locations in and around South East Asia are very cloudy and generally have a very high noise floor which makes manual identification of plumes difficult. 
In Australia, false positives often appear over quarries and mining facilities and as such enhancements are easily confused and true events might be missed by the analysts as a consequence.

\subsection{Detection of \nhthree, \notwo and CO point source events}
\label{res_other_gases}

Finally, we show the results of adapting our system to other \fixS{has}\fixM{gas\label{rev1_q2}}{R1-Q2} species (\nhthree, \notwo and CO) as described in Section \ref{method_spectral_scoring}. 
Before this work only a few isolated events of \nhthree and \notwo emissions had been found in data from VSWIR imaging spectrometers and so far no work reported on CO events.

Adaptation of our system from methane to other gas species requires three distinct steps: (1) computing enhancement products using target signatures of these other gases; (2) zero-shot generalisation of machine learning ensembles pre-trained on a dataset of methane point sources; and (3) using spectral fit metrics adapted to the specific absorption feature of each new gas (in case of \nhthree, using multiple absorption features).


We start with a representative set of over 7.2k EMIT scenes in locations typical for methane point sources as described in \ref{method_data}.
Upon discovery of a new point source event, we process all other EMIT observations over the same location.
Other heuristics can be used to select which scenes to explore first from the catalogue of over \totalEMITscenes (and growing) EMIT scenes currently available.
Notably, detections from other instruments such as TROPOMI could be used to inform future scene selection, as has been done in the context of \chfour in \cite{maasakkers2022using}.

\begin{figure}[ht!]
\setlength\tabcolsep{1pt} 
\centering
\begin{tabular}{@{} l M{0.43\linewidth} M{0.43\linewidth} @{}}
\begin{subfigure}{0.08\linewidth} \caption{\nhthree}\label{subfig:nh3} \end{subfigure} 
  & \includegraphics[width=\hsize]{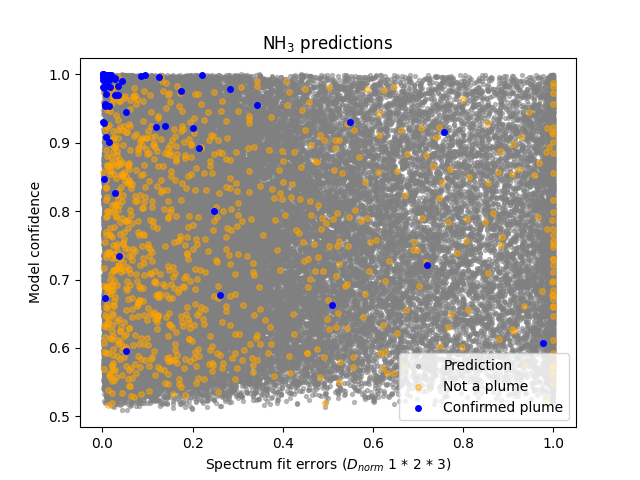} 
  & \includegraphics[width=\hsize]{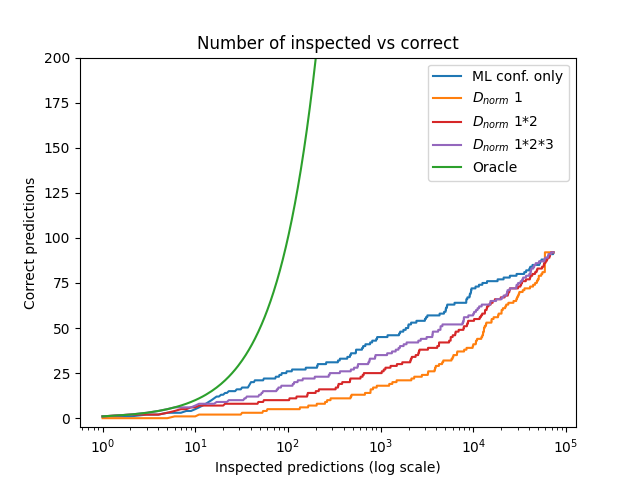}\\ \addlinespace
\begin{subfigure}{0.08\linewidth} \caption{\notwo}\label{subfig:no2} \end{subfigure} 
  & \includegraphics[width=\hsize]{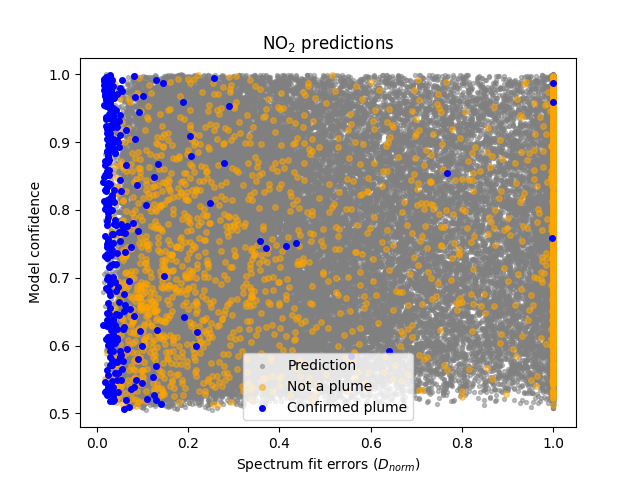}    
  & \includegraphics[width=\hsize]{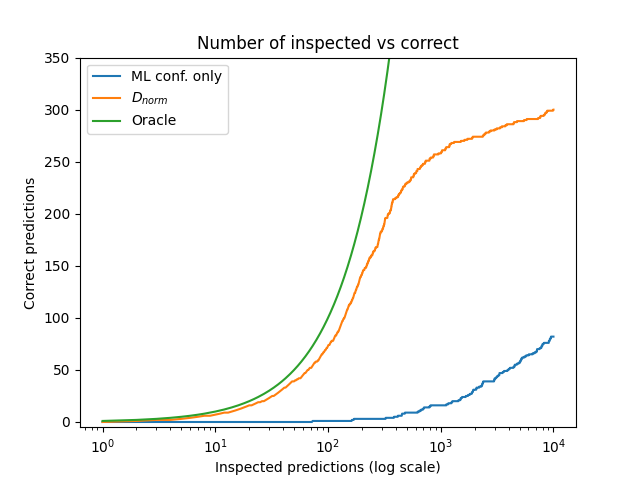}\\ \addlinespace
\begin{subfigure}{0.08\linewidth} \caption{CO}\label{subfig:co} \end{subfigure} 
  & \includegraphics[width=\hsize]{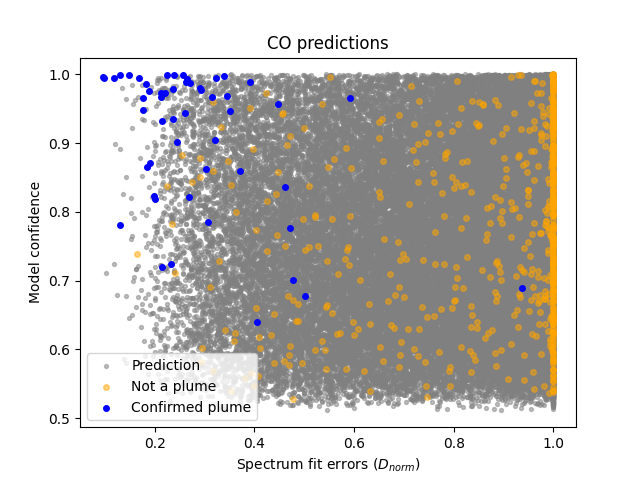} 
  & \includegraphics[width=\hsize]{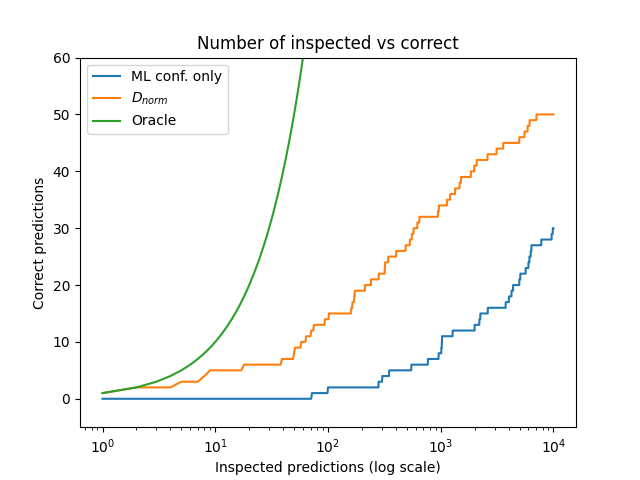}\\
\end{tabular}
\caption{System for detection of \nhthree, \notwo and CO point source emitters. On the left we show scored predictions computed for all 7.2k EMIT scenes highlighting true detected point sources in blue and confirmed non-sources in yellow. For better clarity we show only predictions larger than 100px. On the right, we compare the performance of using different metrics for prioritization of manually inspected predictions. }
\label{fig:scores_prioritizations_all_three}
\end{figure}

Figure \ref{fig:scores_prioritizations_all_three} shows the distribution of spectral fit and model confidence scores over the entire set of EMIT scenes selected for \nhthree, \notwo and CO.
We mark a prediction as correct when it overlaps with the manually delineated ground truth label.
As with methane in Section \ref{res_method_verif}, we can use these scores to prioritize scenes for manual inspection. However in contrast with the spectral fit metrics computed for methane, the scores for \nhthree, \notwo and CO tell a different story.

Particularly in the case of \nhthree in Figure \ref{subfig:nh3}, we find that false alarms and true detections are harder to separate than for methane. 
However, ML model confidences are much more useful than for \chfour (comparing with the plots shown on Figure \ref{fig:res_prioritization}). 
We compute \dnorm scores for multiple different spectral ranges and denote them as \dnorm1 (1498-1603 nm), \dnorm2 (1952-2123 nm) and \dnorm3 (2123-2326 nm). As all three should be low for true point source events, we compute their product and show that using multiple \dnorm values outperforms using just a subset. Surprisingly, however, the best performing prioritization function is the one that uses only ML model confidences.

The \nhthree plume aligned with a known methane plume in 13 scenes. This alignment might be caused either by co-emission (e.g. when \chfour is used as ingredient for fertilizer generation), or by a false positive from spectrally-similar features at 2300 nm. To exclude the second possibility, we computed matched filter variants without the 2 micron spectral region containing overlapping features and removed 5 events. In the remaining 8 cases, the \nhthree plumes remained clearly visible despite having ignored the \chfour channels, so they are likely co-emitted \nhthree and \chfour.

In the case of \notwo in Figure \ref{subfig:no2}, false alarms and true detections have overlapping \dnorm and model confidence metrics. Notably, ML model confidences are not as useful as in the case of \nhthree~- the confirmed true events are spread across the entire ML confidence range.
This is an interesting finding, it may be due to the different shape of the typical \notwo plume, as these tend to be more disperse than either \nhthree or \chfour plumes. 
Models pre-trained on different types of plume shapes may struggle with some of these more dispersed examples.

Finally, for CO point source point sources shown in Figure \ref{subfig:co}, we note that both model confidence and \dnorm scores are useful. However the prioritization plot on the right shows that using \dnorm outperforms ML confidence only mode.

\begin{figure}[!t]
    \centering
    
    \includegraphics[trim=1.5cm 1.5cm 1.5cm 1.5cm,clip,width=0.9\linewidth]{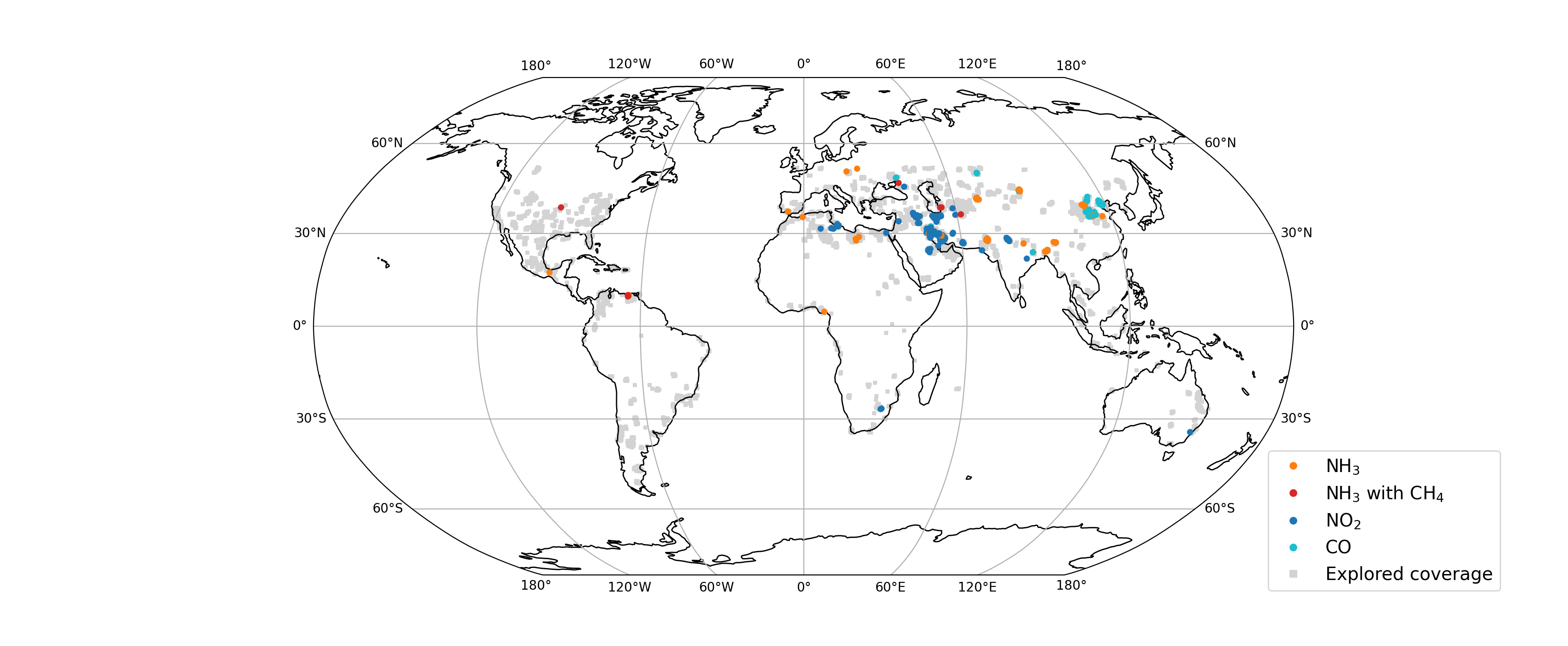}
    \caption{Map of \nhthreeEvents newly found \nhthree, \notwoEvents \notwo and \coEvents CO events in the archive of over 7.2k considered EMIT scenes (coverage of these scenes is shown in light gray). Note that the released dataset also contains a set of confounders and no-plume scenes, for clarity these are not shown here.}
    \label{fig:map_other_gases}
\vspace{-2mm}
\end{figure}

Figure \ref{fig:map_other_gases} shows the global distribution of the newly found \nhthree, \notwo and CO point sources. 
Supplementary materials show selected examples of discovered events and found confounder scenes.
Each detected source has been co-located with a industrial site - power plants for \cotwo, usually ammonia processing sites for \nhthree and steel processing sites for CO.

\subsection{Limitations}

While the system described in this paper is fully automatic, our experiments still required manual analyst verification of alerts produced by the system. 
However, this manual verification is only required to analyze the performance of the system to ascertain that it could be deployed in fully automated manner.
With the performance showcased in this paper, and the operational deployment of the Daily Methane Digest, full automation is feasible and has been achieved for the detection of methane point sources.
We discover that for other trace gases like \notwo, \nhthree and CO, our system does not reach the same capabilities as for \chfour.  This is mainly due to the different morphological properties, such as the diffuse plumes of \notwo, and weaker overall absorption features of the other gases. 
This challenge is an inherent limitation of the current generation of imaging spectroscopy instruments and may improve with their future variants having improved radiometric precision and/or finer spectral sampling.
Additionally, the machine learning models used were primarily trained on methane sources, and their generalisation may suffer when expanded to additional trace gases.

The system as presented meets the challenging criteria of an extremely low false positive rate.  Doing so, even with the hybrid physics and machine learning strategy leveraged, means accepting a lower true positive rate - in other words smaller (less obvious) methane sources may be missed.  However, it is the large methane sources which are most important for timely detection and mitigation.
Our experiments also show that at the cost of additional manual labour, even smaller methane point sources can be detected.  Here, the prioritization schema generated by our system improves accuracy and efficiency over a purely manual assessment.

\section{\fix{Conclusion}}
\fixM{\label{rev2_q7}}{R2-Q7}

In conclusion, this paper presents a new system the for detection of trace gas plume events in imaging spectroscopy data. It demonstrates the combination of a machine learning model for plume delineation with physics-based spectroscopy for robust classification, effectively combining the best of both visual pattern recognition with spectroscopic modelling.
We utilize this system to detect not only signatures previously identified from imaging spectroscopy data such as \chfour, \nhthree, and \notwo, but also the novel signature of CO.

We report several specific use cases for this system. 
First, we demonstrate fully automated global and daily monitoring of large methane leak events. Our Daily Methane Digest has been operational since January 2026, and in the \monthsRunningDaily initial months of deployment detected over \dailyFoundEvents methane leak events which were directly fed into the review process described in \cite{emit_plumes_v002_atbd} and eventually will be published as data products in \cite{emit_plumes_v002_data_product}.
Importantly, the estimated time-to-information has been dramatically reduced, from more than a month to near same-day latency. 
In a recent large methane leak, an alert from this system was used to schedule an observation with the Planet’s Tanager-1 satellite, which confirmed that the emission has ceased. 

Second, we use our system to search through two full years of EMIT data finding high confidence methane leak events which were previously missing from methane catalogues of \cite{emit_plumes_v002_data_product}.
Filling the gaps of knowledge about methane leak events is crucial for decision making about future detected events - as repeated detections over the same location serve as further evidence and are easier to verify. Temporal statistics about methane leak emissions can also inform us about the consistency of mitigation cases. The automated approach has the potential to be less biased and more systematic than manual plume identification.

We adapt this system to detection of trace gases other than methane. 
Our system leverages the common morphological structure of different trace gas plume emissions to transfer a machine learning model trained on \chfour plumes to these additional gas species.  When combined with the physics-based filtering criteria, this translates to an effective scaling solution. We were also able to utilize this system to investigate trace gases previously not published in the literature, and in so doing identify CO point source emissions in the EMIT data.
We publish new datasets with in total \nhthreeEvents \nhthree, \notwoEvents \notwo \coEvents CO point source events. These datasets follow best practice standards and are complemented with selected scenes with common confounder features and matching number of background non-emission scenes.

In the future, there are numerous potential research directions and extensions to consider. The combination of matched filter preprocessing with machine learning model and spectral fitting is modular, permitting incremental improvements to any of the three steps independently. 
Future variants of matched filter products could create better enhancement fields that improve the separation of the events from the background and confounder features.
For example, methods like that of \cite{fahlen2024sensitivity} could compensate for simplifications in the matched filter formulation that leave it less sensitive over dark surfaces. 
Future machine learning models could also leverage information from the imaging spectroscopy data directly by using more spectral bands as inputs. Initial work in this area was demonstrated for methane leak detection in \cite{HyperspectralViTs}, \cite{kumar2023methanemapper} and in the recent preprint of \cite{batchu2026preprint} - other tasks such as cloud detection were explored in \cite{lee2025spectf}.
We also highlight that there are new research directions enabled by our released datasets of trace gases. These could focus on learning methods in regimes with limited labels such as semi-supervised or unsupervised learning. These datasets could also be used as benchmark evaluation targets of future hyperspectral foundational models.

Finally, we aim to keep running the Methane Daily Digest as an open public platform and to extend it to monitoring of \nhthree, \notwo and CO point sources in daily EMIT data. We will also deploy our method on the full archive of EMIT data, as there are likely events of interest we have missed by only exploring the initial (although large) set of locations.

\bibliography{bib}

\section*{Funding}

This work was carried out at the Jet Propulsion Laboratory, California Institute of Technology, under a contract with the National Aeronautics and Space Administration (80NM0018D0004). Part of Vít Růžička's research was supported by an appointment to the NASA Postdoctoral Program at the Jet Propulsion Laboratory, administered by Oak Ridge Associated Universities under contract with NASA.

\section*{Author contributions statement}

VR is responsible as the first and corresponding author for Conceptualization, Methodology, Software, Analysis, Writing and Expirements. DRT for Conceptualization, Methodology and Writing. JF has generated spectral targets. AMT, SL, HB, DJ and VR were responsible for Data Curation. AT for Analysis and Interpretation. ROG, AT and DRT for Supervision. All authors reviewed and contributed to the manuscript.

\section*{Data availability statement}

All used EMIT radiance data is available through the NASA's Earthdata Search (\url{https://www.earthdata.nasa.gov/data/catalog/lpcloud-emitl1brad-001}). 
We release datasets of newly found \nhthree, \notwo and CO events at \zenodoDataset. 
Finally, the codebase including the weights of the trained machine learning models is available at \codeModels. An included jupyter notebook will download any EMIT scene ID and run the model segmentation and scoring.

\section*{Competing interests}

The authors declare no competing interests.

\newpage

\begin{center}
  \makebox[\textwidth][c]{\bfseries\Large Supplementary Materials for Fully Automatic Trace Gas Plume Detection}
\end{center}

\begin{center}

Vít~Růžička$^{\ast}$,
David~R.~Thompson,
Jay~E.~Fahlen,
Amanda~M.~Lopez,
Steven~Lu,
Chuchu~Xiang,
Holly~Bender,
Daniel~Jensen,
Phil~Brodrick,
Jake~Lee,
Brian~Bue,
Daniel~H.~Cusworth,
Luis~Guanter,
Adam~Chlus,
Andrew~Thorpe,
Robert~O.~Green\\
\small$^\ast$Corresponding author. Email: ruzicka@jpl.nasa.gov\\
\end{center}

\setcounter{figure}{0}
\setcounter{table}{0}
\renewcommand{\thefigure}{S\arabic{figure}} 
\renewcommand{\thetable}{S\arabic{table}} 


\subsection*{Text S1. Adapting spectral fitting to other trace gases}

We extend the plume transmittance methods of \cite{xiang2025identification} for trace gases other than methane, selecting different spectral ranges according to the absorption features of each target. Figure \ref{fig:gas_species_targets_ranges} details this modification. We first calculate column-wise matched filter (CMF) matched filter products using the target signatures of these other trace gas species (Figure \ref{fig:example_trace_gas_plumes}). 

Second, we apply the machine learning models described in Section \ref{method_semseg} to detect of plume-like shapes in these matched filter products.  The models are applied in zero-shot fashion without any adaptation, trained on a dataset of methane sources with just the \chfour-CMF on its input.  Since all gases function as passive tracers in the atmosphere, we hypothesize that the plume shapes will in principle be the same for both.  In practice, of course, aspects of the CMF product might differ slightly for the different spectral signatures, including variable influence of surface albedo and background clutter.  

Third, as with methane, we score each predicted plume with spectral fit metrics adapted to each specific trace gas. The overview of the hyperparameters used for these gases is listed in Table \ref{tab:plume_vetting_params}. Parameters not listed in this table are kept the same default value for all species. We compare the plume transmittance spectrum to the known target gas absorptions. In each interval, we fit a spectral model based on a polynomial continuum perturbed by some amount of Beer-Lambert absorption by the target species.  The polynomial represents any residual difference in surface reflectance that was not removed by the background ratio.  The Beer-Lambert absorption gives an estimate of the optical depth.  For methane, which has lines that overlap water absorptions in the shortwave infrared, we calculate the effective absorption coefficients at high resolution with an atmosphere and observing geometry appropriate for the scene.  For the other target gases, we use absorption coefficients from the MODTRAN line list. We fit a combined model for the plume transmittance including all polynomial coefficients with the plume optical depth. 

\begin{table}[h]
\caption{Parameters for plume vetting. Note that in case of \nhthree we obtain three \dnorm scores according to the used in-band range. MF threshold is used to select low enhancement pixels during background sampling. }
\label{tab:plume_vetting_params}
\centering
\scalebox{1.0}{
\begin{tabular}{@{}llll@{}}
\toprule
 & In-band ranges (nm) & Out-band ranges (nm) & MF thr. \\ \midrule
\chfour & \textless{}2100; 2440\textgreater{} & \textless{}381; 1633\textgreater{}, \textless{}1692; 2094\textgreater{}, \textless{}2441; 2493\textgreater{} & 30 \\ \midrule
\nhthree & \begin{tabular}[c]{@{}l@{}}1: \textless{}1498; 1603\textgreater\\ 2: \textless{}1952; 2130\textgreater\\ 3: \textless{}2123; 2326\textgreater{}\end{tabular} & \textless{}381; 1498\textgreater{}, \textless{}1603; 1922\textgreater{}, \textless{}2441; 2493\textgreater{} & 300 \\ \midrule
\notwo & \textless{}381; 753\textgreater{} & \textless{}753; 1633\textgreater{}, \textless{}1692; 2094\textgreater{}, \textless{}2441; 2493\textgreater{} & 1200 \\  \midrule
CO & \textless{}2278; 2441\textgreater{} & \textless{}381; 1633\textgreater{}, \textless{}1692; 2094\textgreater{}, \textless{}2441; 2493\textgreater{} & 300 \\ \bottomrule
\end{tabular}
}
\end{table}



\subsection*{Text S2. Example \nhthree, \notwo and CO events}

\begin{figure}[!h]
    \centering
    \includegraphics[trim=0 4.4cm 0 0cm,clip,width=0.7\textwidth]{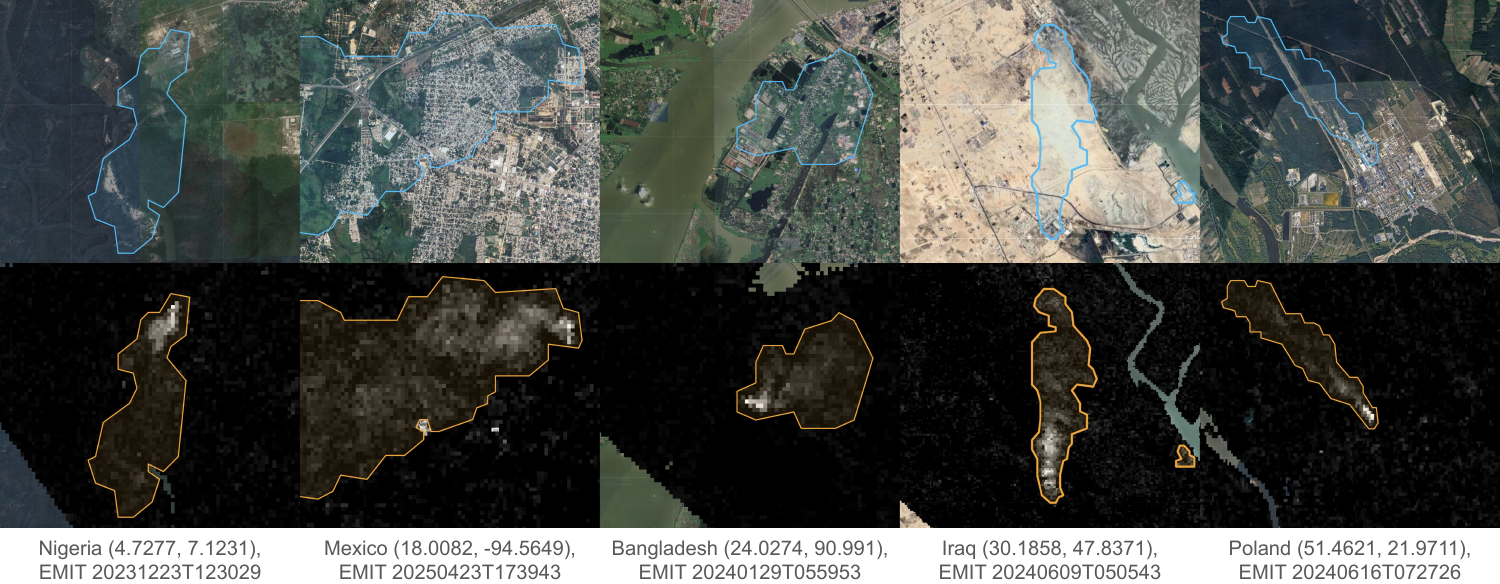}
    \caption{Example \nhthree point sources detected around the world in EMIT data. 
    We highlight the 2nd and 3rd examples, which occur in densely populated regions. From left to right, emission sources include a fertilizer plant in Nigeria, ammonia plant in Mexico, fertilizer plant in Bangladesh, petrochemical plant in Iraq, and urea plant in Poland.}
    \label{fig:nh3_events}
\vspace{-2mm}
\end{figure}

\begin{figure}[!h]
    \centering
    
    \includegraphics[trim=0 4.4cm 0 0cm,clip,width=0.7\textwidth]{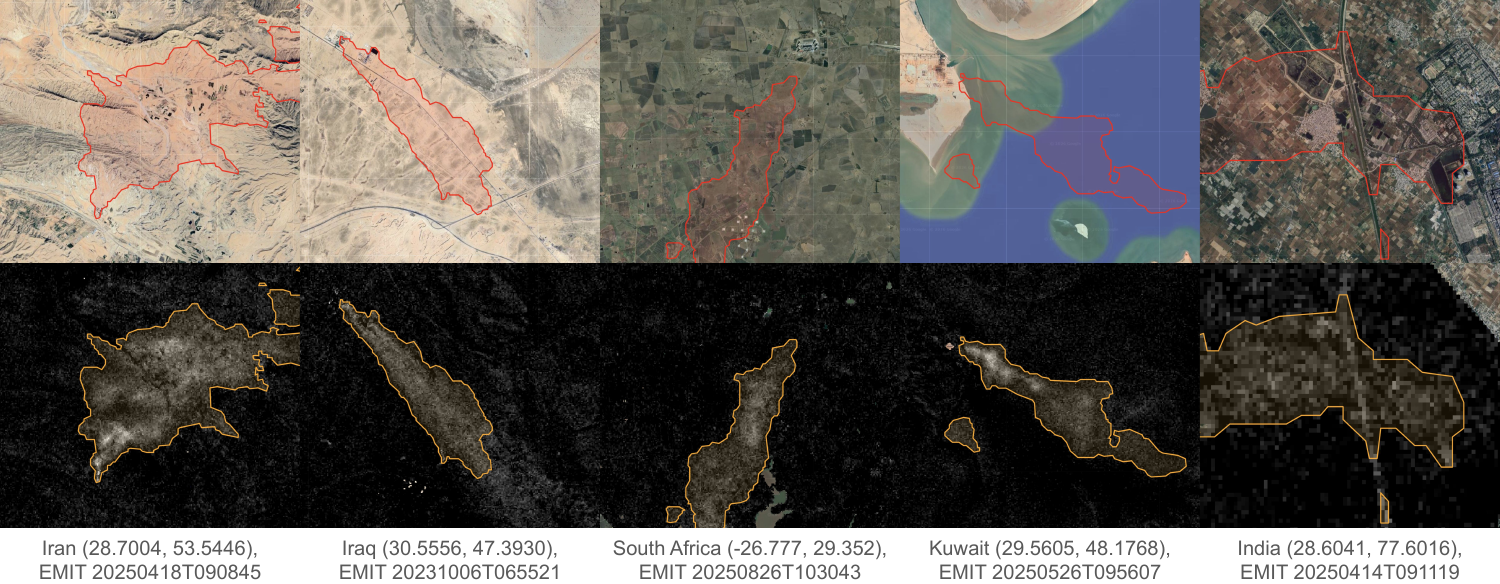}
    \caption{Example \notwo point sources detected around the world in EMIT data. 
    All emission sources are from power plants.}
    \label{fig:no2_events}
\vspace{-2mm}
\end{figure}

\begin{figure}[!h]
    \centering
    \includegraphics[trim=0 4.4cm 0 0cm,clip,width=0.7\textwidth]{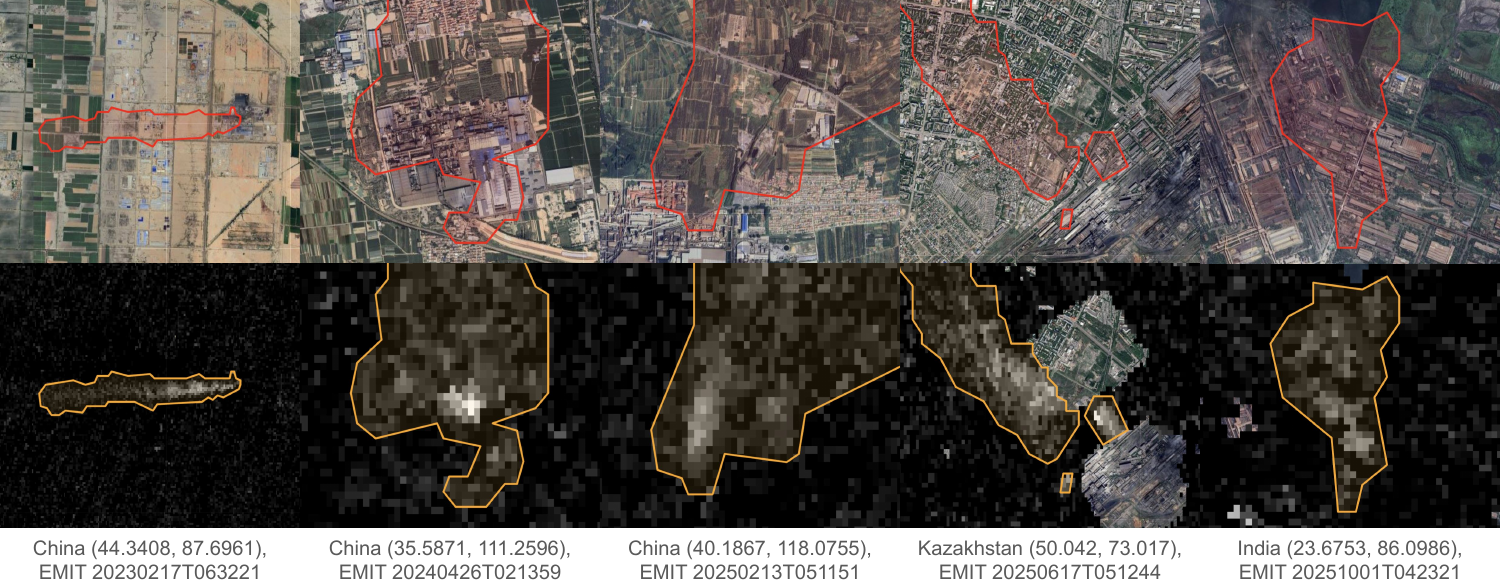}
    \caption{Example CO point sources detected around the world in EMIT data. 
    From left to right, emission sources include two chemical plants in China, possible cement production in China, and steel plants in Kazakhstan and India.}
    \label{fig:co_events}
\vspace{-2mm}
\end{figure}

Selected examples for \nhthree are on Figure \ref{fig:nh3_events}, \notwo on Figure \ref{fig:no2_events} and finally for CO events on Figure \ref{fig:co_events}.
Top rows shows the location of the found events on high resolution satellite imagery provided through Google Earth. On the bottom row, we show the computed enhancement product with the vector predicted by our machine learning models. Names of the used EMIT scenes can be uniquely identified by the provided capture time.
Particularly striking are the detections of gas plumes over populated areas (2nd and 3rd example in \nhthree and last example in \notwo).

\subsection*{Text S3. Exploring confounder events for \nhthree, \notwo and CO}

\begin{figure}[ht]
    \centering
    \begin{subfigure}[b]{0.7\textwidth}
        \centering
        \includegraphics[trim=0 1.5cm 0 0cm,clip,width=0.9\textwidth]{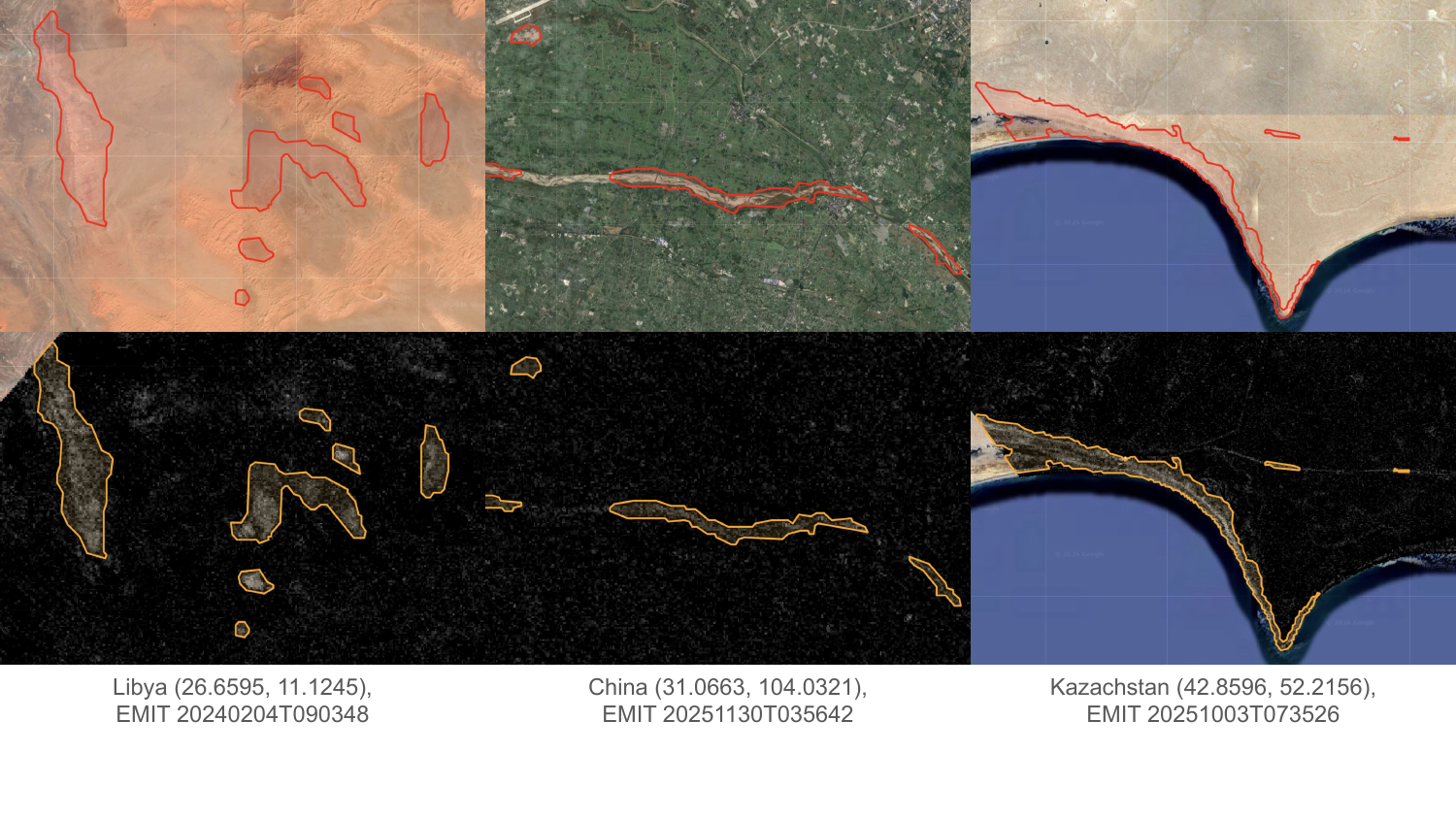}
        \caption{Example \nhthree confounders.}
    \end{subfigure}
    \vspace{0.3cm}

    \begin{subfigure}[b]{0.7\textwidth}
        \centering
        \includegraphics[trim=0 1.5cm 0 0cm,clip,width=0.9\textwidth]{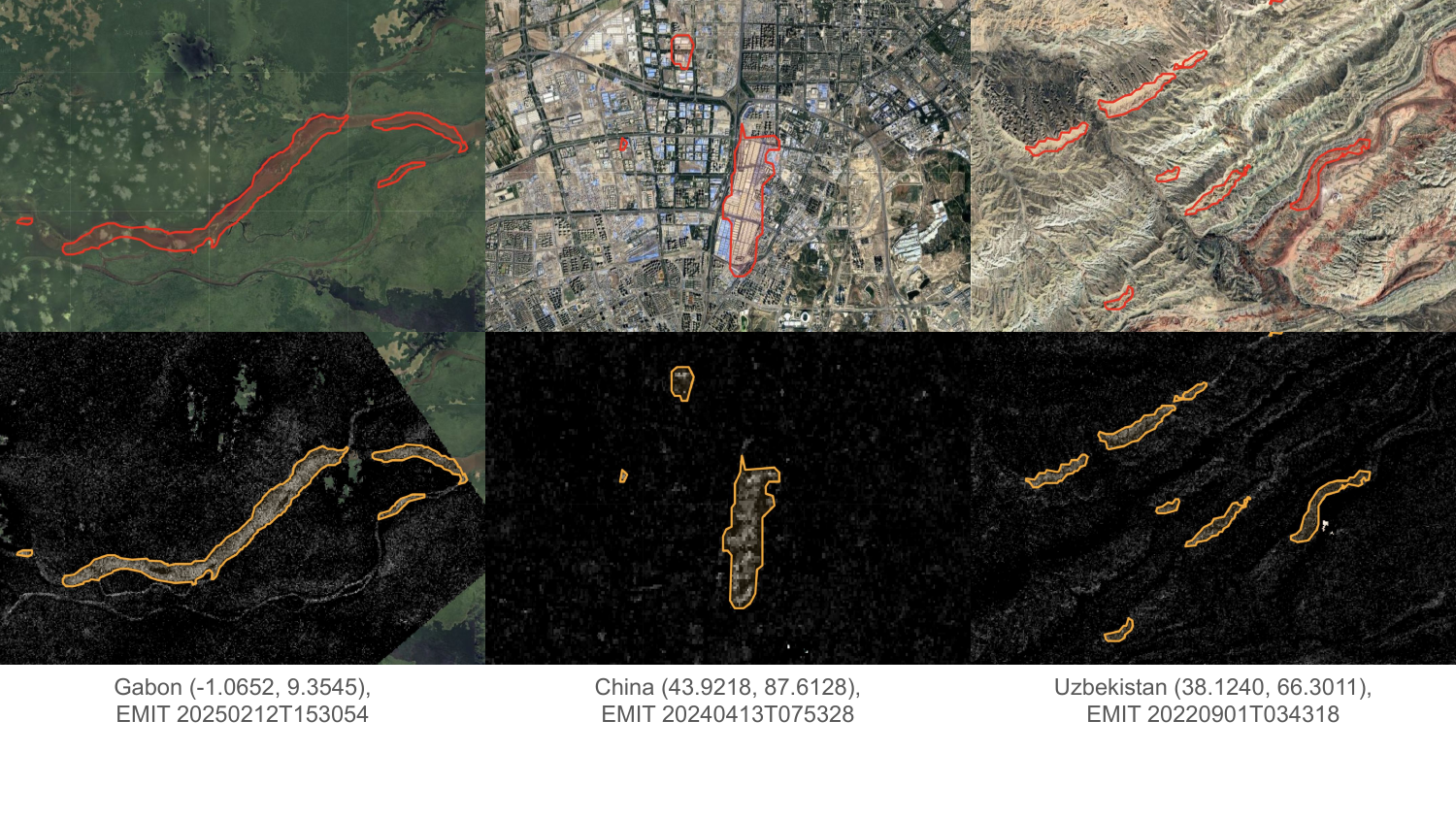}
        \caption{Example \notwo confounders.}
    \end{subfigure}
    \vspace{0.3cm}

    \begin{subfigure}[b]{0.7\textwidth}
        \centering
        \includegraphics[trim=0 1.5cm 0 0cm,clip,width=0.9\textwidth]{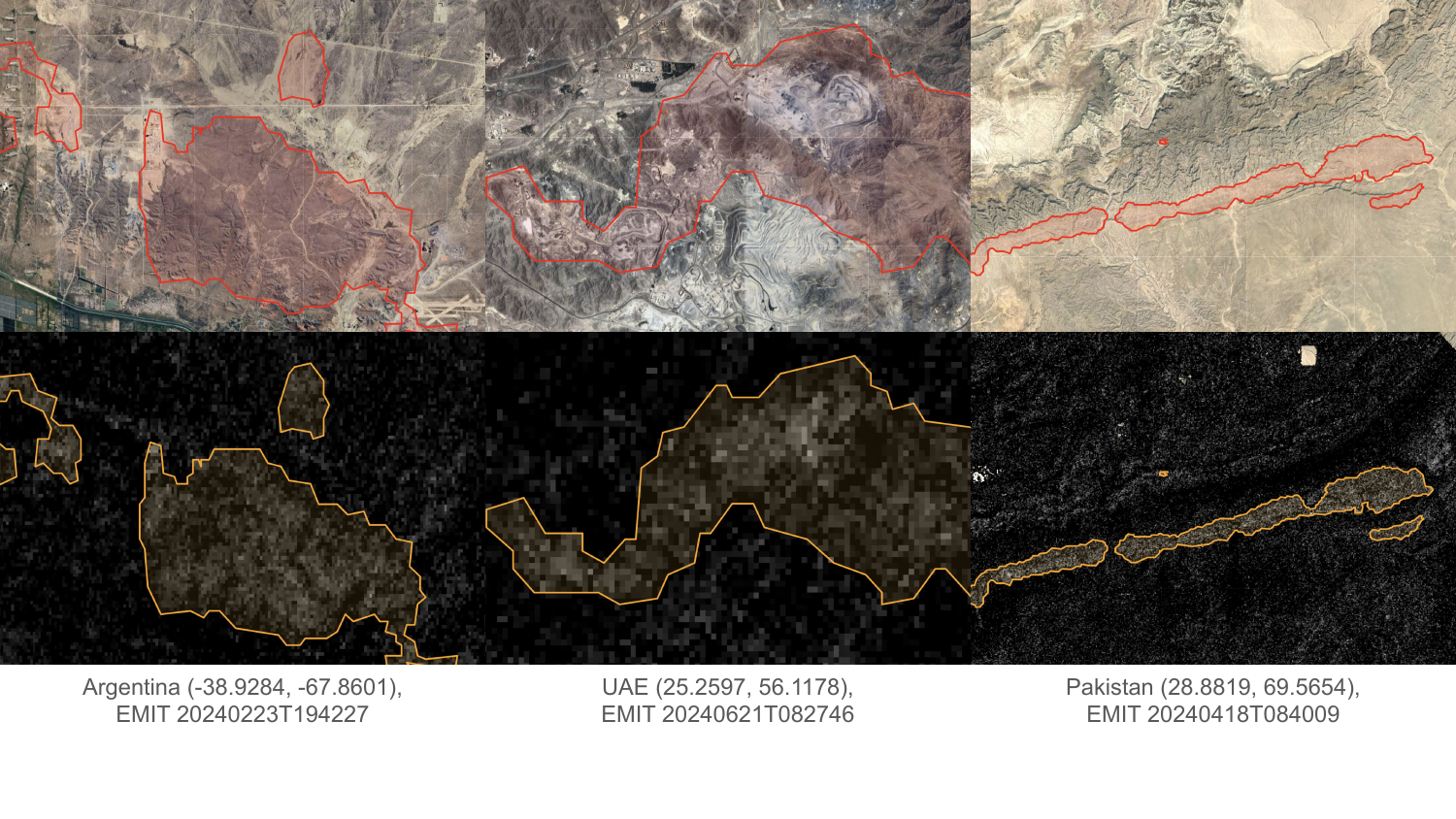}
        \caption{Example CO confounders.}
    \end{subfigure}

    \caption{Example \nhthree, \notwo and CO confounders. While in the case of \nhthree and CO most of these are primarily mineralogical, in the case of \notwo we observe potential biological triggers. Names of the used EMIT scenes can be uniquely identified by the provided capture time.}
    \label{fig:confounders}
\vspace{-2mm}
\end{figure}

In Figure \ref{fig:confounders} we show examples of new types of confounders found for \nhthree, \notwo and CO. 
We highlight that the spectral fit metrics introduced in this paper are effictive for disambiguation between confounder and true point source events.

We believe these examples can be useful to guide future research directions - for example many of \notwo confounders occur over water bodies, river deltas and oxbows. This could be of potential biological origins, for example as an indication of the chlorophyll \textit{a} (which has a strong absorption feature around \fix{440 nm}) and algae or phytoplankton \cite{lee2004absorptionChlorophyll}.
Computing water masks and filtering out detections over water bodies could be one easy way of addressing these confounders. 
Certain mineral species also trigger matched filter products for \nhthree and \notwo (as is also the case of \chfour).
In case of CO confounders, lime stone mines (shown on first two scenes) may be triggering the MF enhancement, likely due to the absorption feature of calcium carbonate (CaCO$_3$) around 2300 nm \cite{cao2024semi_CaCO3}.

\clearpage 

\end{document}